# Lifted Variable Elimination:
# Decoupling the Operators from the Constraint Language

**Nima Taghipour**                                   NIMA.TAGHIPOUR@CS.KULEUVEN.BE
**Daan Fierens**                                      DAAN.FIERENS@CS.KULEUVEN.BE
**Jesse Davis**                                       JESSE.DAVIS@CS.KULEUVEN.BE
**Hendrik Blockeel**                                  HENDRIK.BLOCKEEL@CS.KULEUVEN.BE
*KU Leuven, Department of Computer Science*
*Celestijnenlaan 200A, 3001 Leuven, Belgium*

## Abstract

Lifted probabilistic inference algorithms exploit regularities in the structure of graphical models to perform inference more efficiently. More specifically, they identify groups of interchangeable variables and perform inference once per group, as opposed to once per variable. The groups are defined by means of constraints, so the flexibility of the grouping is determined by the expressivity of the constraint language. Existing approaches for exact lifted inference use specific languages for (in)equality constraints, which often have limited expressivity. In this article, we decouple lifted inference from the constraint language. We define operators for lifted inference in terms of relational algebra operators, so that they operate on the semantic level (the constraints' extension) rather than on the syntactic level, making them language-independent. As a result, lifted inference can be performed using more powerful constraint languages, which provide more opportunities for lifting. We empirically demonstrate that this can improve inference efficiency by orders of magnitude, allowing exact inference where until now only approximate inference was feasible.

## 1. Introduction

Statistical relational learning or SRL (Getoor & Taskar, 2007; De Raedt, Frasconi, Kersting, & Muggleton, 2008) focuses on combining first-order logic with probabilistic graphical models, which permits algorithms to reason about complex, uncertain, structured domains. A major challenge in this area is how to perform inference efficiently. First-order logic can reason on the level of logical variables: if a model states that for all $X$, $P(X)$ implies $Q(X)$, then whenever $P(X)$ is known to be true, one can infer $Q(X)$, without knowing what $X$ stands for. Many approaches to SRL, however, transform their knowledge into a propositional graphical model before performing inference. By doing so, they lose the capacity to reason on the level of logical variables: standard inference methods for graphical models can reason only on the "ground" level, repeating the same inference steps for each different value $x$ of $X$, instead of once for all $x$.

To address this problem, Poole (2003) introduced the concept of lifted inference for graphical models. The idea is to group together interchangeable objects, and perform the inference operations once for each group instead of once for each object. Multiple different algorithms have been proposed, based on variable elimination (Poole, 2003; de Salvo Braz, Amir, & Roth, 2005; Milch, Zettlemoyer, Kersting, Haimes, & Kaelbling, 2008; Sen, Deshpande, & Getoor, 2009b, 2009a; Choi, Hill, & Amir, 2010; Apsel & Brafman, 2011), belief propagation (Kersting, Ahmadi, & Natarajan, 2009; Singla & Domingos, 2008), or various other approaches (Van den Broeck, Taghipour, Meert, Davis, & De Raedt, 2011; Jha, Gogate, Meliou, & Suciu, 2010; Gogate & Domingos, 2011).

A group of interchangeable objects is typically defined by means of a constraint that an object must fulfill in order to belong to that group. The type of constraints that are allowed, and the way in which they are handled, directly influence the granularity of the grouping, and hence, the efficiency of the subsequent lifted inference (Kisynski & Poole, 2009a). Among the approaches based on variable elimination, the most advanced system (C-FOVE) uses a specific class of constraints, namely, conjunctions of pairwise (in)equalities. This is the bare minimum required to be able to perform lifted inference. However, as we will show, it unnecessarily limits the symmetries the model can capture and exploit.

In this article, we present an algorithm for lifted variable elimination that is based on C-FOVE, but uses a constraint language that is extensionally complete, that is, for any group of variables a constraint exists that defines exactly that group. To this aim, C-FOVE's constraint manipulation is redefined in terms of relational algebra operators. This decouples the lifted inference algorithm from the constraint representation mechanism. Consequently, any constraint language that is closed under these operators can be plugged into the algorithm to obtain a working system. Apart from redefining existing operators, we also define a novel operator, called lifted absorption, in this way. Furthermore, we propose a concrete mechanism for constraint representation that is extensionally complete, and briefly discuss how the operators can be implemented for this particular mechanism. The new lifted inference algorithm, with this constraint representation mechanism, can perform lifted inference with a much coarser granularity than its predecessors. Due to this, it outperforms existing systems by several orders of magnitude on some problems, and solves inference problems that until now could only be solved by approximate inference methods.

The basic ideas behind our approach have been explained in an earlier conference paper (Taghipour, Fierens, Davis, & Blockeel, 2012). This article extends that paper by providing precise and motivated definitions for the operators, up to the level where they can be implemented. These definitions, at the same time, may help understand on a more intuitive and semantic level how lifted variable elimination works, and can serve as a kind of gold standard for other implementations of lifted variable elimination, as they provide a semantics based reference point.

The paper is structured as follows. Section 2 illustrates the principles of lifted variable elimination by example, and briefly states how this work improves upon the state of the art, C-FOVE (Milch et al., 2008). Section 3 introduces formal notation and terminology, and Section 4 provides a high-level outline of our lifted variable elimination algorithm. Section 5 describes in detail all the operators that the algorithm uses. Section 6 briefly discusses an efficient representation for the constraints themselves. Section 7 empirically compares our system's performance with that of C-FOVE, and Section 8 concludes.

## 2. Lifted Variable Elimination by Example

Although lifted variable elimination builds on simple intuitions, it is relatively complicated, and an accurate description of it requires a level of technical detail that is not conducive to a clear understanding. For this reason, we first illustrate the basic principles of lifted inference on a simple example, and without referring to the technical terminology that is introduced later. We start with describing the example; next, we illustrate variable elimination on this example, and show how it can be lifted.

## 2.1 The Workshop Example

This example is from Milch et al. (2008). Suppose a new workshop is organized. If the workshop is popular (that is, many people attend), it may be the start of a series. Whether a person is likely to attend depends on the topic.

We introduce a random variable $T$, indicating the topic of the workshop, and a random variable $S$, indicating whether the workshop becomes a series. We consider $N$ people, and for each person $i$, we include a random variable $A_i$ that indicates whether $i$ attends. Each random variable has a finite domain from which it takes on values, i.e., $\{ai, ml, \dots\}$ for $T$, $\{yes, no\}$ for $S$, and $\{true, false\}$ for each $A_i$.

The joint probability distribution of these variables can be specified by an undirected graphical model. A set of *factors* captures dependencies between the random variables in such a model. In our model, there are two kinds of factors. For each person $i$, there is a factor $\phi_1(A_i, S)$ that states how having a series depends on whether person $i$ attends, and a factor $\phi_2(T, A_i)$ that states how $i$'s attendance depends on the topic. Note that all $N$ factors of the first type have the same potential function $\phi_1$, and all factors of the second type have potential function $\phi_2$. This imposes a certain symmetry on the model: it implies that $S$ depends on each person's attendance in exactly the same way, and all people have the same topic preferences.

The model defines a joint probability distribution over the variables that is the normalized product of the factors (normalized such that all joint probabilities sum to one):

$$Pr(T, S, A_1, \dots, A_N) = \frac{1}{Z} \prod_{i=1}^{n} \phi_1(A_i, S) \prod_{i=1}^{n} \phi_2(T, A_i)$$

where $Z$ is the normalization constant.

Undirected graphical models can be visualized as *factor graphs* (Kschischang, Frey, & Loeliger, 2001), which have a node for each random variable and each factor, and an edge between a factor and a random variable if that variable occurs in the factor. Figure 1 shows a factor graph for our example.

## 2.2 Variable Elimination

From now on, we refer to the values taken by a variable by the corresponding lowercase symbols (e.g., $a_i$ as shorthand for $A_i = a_i$).

Suppose we want to compute the marginal probability distribution $Pr(S)$.

$$Pr(S) = \sum_T \sum_{A_1} \cdots \sum_{A_N} Pr(T, S, A_1, \dots, A_N) \tag{1}$$

$$= \frac{1}{Z} \sum_T \sum_{A_1} \cdots \sum_{A_N} \prod_{i=1}^{N} \phi_1(A_i, S) \prod_{i=1}^{N} \phi_2(T, A_i) \tag{2}$$

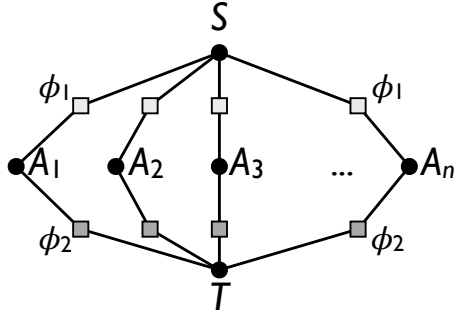

Figure 1: A factor graph for the workshop example. Square nodes represent factors, round nodes variables. Variables are labeled with their name, factors with their potential function.

Usually, the normalization constant $Z$ is ignored during the computations, and normalization happens only at the very end. So, we can focus on how to compute

$$\tilde{Pr}(S) = \sum_T \sum_{A_1} \cdots \sum_{A_N} \prod_{i=1}^N \phi_1(A_i, S) \prod_{i=1}^N \phi_2(T, A_i). \qquad (3)$$

A straightforward way of computing $\tilde{Pr}(S)$ is to compute $\tilde{Pr}(s)$ for each possible value $s$ of $S$, and tabulate the results. We can compute $\tilde{Pr}(true)$ by iterating over all possible value combinations $(t, a_1, \ldots, a_n)$ of $(T, A_1, \ldots, A_n)$ and computing $\prod_{i=1}^N \phi_1(a_i, true) \prod_{i=1}^N \phi_2(t, a_i)$ for each combination, and similarly for $\tilde{Pr}(false)$. If all variables are binary, there are $2^{N+1}$ such combinations, and for each combination $2N - 1$ multiplications are performed. This clearly does not scale.

However, we can improve efficiency by rearranging the computations. In the above computation, the same multiplications are performed repeatedly. Since $\phi_1(A_1, S)$ and $\phi_2(T, A_1)$ are constant in all $A_i$ except $A_1$, they can be moved out of the summations over $A_i$, $i > 1$, so the right hand side of Equation 3 becomes:

$$\sum_T \sum_{A_1} \phi_1(A_1, S) \phi_2(T, A_1) \sum_{A_2} \cdots \sum_{A_N} \prod_{i=2}^N \phi_1(A_i, S) \prod_{i=2}^N \phi_2(T, A_i) \qquad (4)$$

Conversely, the factor starting with $\sum_{A_2}$ is independent of $A_1$, so it can be moved outside of the summation over $A_1$, giving

$$\sum_T \left( \sum_{A_2} \cdots \sum_{A_N} \prod_{i=2}^N \phi_1(A_i, S) \prod_{i=2}^N \phi_2(T, A_i) \right) \left( \sum_{A_1} \phi_1(A_1, S) \phi_2(T, A_1) \right) \qquad (5)$$

Repeating this for each $A_i$ eventually yields

$$\sum_T \left( \sum_{A_1} \phi_1(A_1, S) \phi_2(T, A_1) \right) \ldots \left( \sum_{A_N} \phi_1(A_N, S) \phi_2(T, A_N) \right) \qquad (6)$$

$$\phi_1(A_1, S) \qquad\qquad \phi_2(T, A_1) \qquad\qquad \phi_{12}(T, A_1, S)$$

| $A_1$ | $S$ | $\phi_1$ |
|-------|------|----------|
| true | true | 1 |
| false | true | 2 |
| true | false | 2 |
| false | false | 1 |

$\otimes$

| $T$ | $A_1$ | $\phi_2$ |
|------|-------|----------|
| SRL | true | 3 |
| SRL | false | 1 |
| DB | true | 2 |
| DB | false | 2 |

$=$

| $T$ | $A_1$ | $S$ | $\phi_{12}$ |
|------|-------|------|-------------|
| SRL | true | true | 3 |
| SRL | true | false | 6 |
| SRL | false | true | 2 |
| SRL | false | false | 1 |
| DB | true | true | 2 |
| DB | true | false | 4 |
| DB | false | true | 4 |
| DB | false | false | 2 |

$$\phi'_{12}(T, S)$$

$\sum_{A_1} \phi_{12}(T, A_1, S) \quad = $

| $T$ | $S$ | $\phi'_{12}$ |
|------|------|--------------|
| SRL | true | 5 |
| SRL | false | 7 |
| DB | true | 6 |
| DB | false | 6 |

Figure 2: Two example factors, their product, and the result of summing out $A_1$ from the product. The values of $\phi_1$ and $\phi_2$ are chosen arbitrarily for this illustration.

which shows that for each $A_i$, the product $\phi_1(A_i, S)\phi_2(T, A_i)$ needs to be computed only once for each combination of values for $(T, S, A_i)$. When $T$ is binary, there are eight such combinations, reducing the total number of multiplications to $8N$.

Note that the result of Formula 6 is a function of $S$; it can yield a different value for each value $s$ of $S$. In other words, it is a factor over $S$. Similarly, the result of $\phi_1(A_1, S) \cdot \phi_2(T, A_1)$ depends on the values of $S$, $T$ and $A_1$ (is a factor over these variables), and after summation over $A_1$ a factor over $S$ and $T$ is obtained. Thus, the multiplications and summations in Formula 6 are best seen as operating on factors, not individual numbers. Figure 2 illustrates the process of multiplying and summing factors.

The result of Formula 6 can be computed as follows. First, multiply the factors $\phi_1(A_1, S)$ and $\phi_2(T, A_1)$ for each value of $A_1$, and sum out $A_1$ from the product. This is exactly the computation illustrated in Figure 2. After summing over all values of $A_1$, the result depends on $T$ and $S$ only; $A_1$ no longer occurs in this factor, nor in any other factors. We say that $A_1$ has been eliminated. Note that the elimination consisted of first gathering all factors containing $A_1$, multiplying them, then summing over all possible values of $A_1$.

After eliminating $A_1$, we can repeat the process for all other $A_i$, each time obtaining a factor over $T$ and $S$. All those factors can then be multiplied and the result summed over $T$, at which point a single factor over $S$ is obtained. This factor equals $\tilde{Pr}(S)$.

The above computation is exactly what Variable Elimination (VE) does. Generally, VE works as follows. It considers one variable at a time, in an order called the *elimination order*. For each considered variable $X$, VE first retrieves all factors that involve $X$, then multiplies these factors together into a single joint factor, and finally sums out $X$, thereby eliminating $X$ from the factor. Hence, in each step, the number of remaining variables strictly decreases (by 1) and also the number of factors decreases (because the set of factors involving $X$ is replaced by a single factor).

The elimination order can heavily influence runtime. Unfortunately, finding the optimal order is NP-hard. In the above example, the elimination order was $A_1, A_2, \ldots, A_N, T$, and this resulted in a computation with $8N$ multiplications, which is $O(N)$.

## 2.3 Lifted Inference: Exploiting Symmetries Among Factors

In the above example, by avoiding many redundant computations, VE obtained an exponential speedup compared to the naive computation discussed before, reducing computation time from $O(2^N)$ to $O(N)$. $N$ can still be large. Even more efficiency can be gained when we know that certain factors have the same potential function.

In our example, VE computes the same product $N$ times: in Expression 6, factors $\phi_1(A_i, S)$ and $\phi_2(T, A_i)$ are the same for all $i$, and so is their product $\phi_{12}(A_i, S, T) = \phi_1(A_i, S)\phi_2(T, A_i)$. It also computes the sum $\sum_{A_i} \phi_{12}(A_i, S, T)$ $N$ times. This redundancy arises because in our probabilistic model all $N$ people behave in the same way, i.e., all $A_i$ are interchangeable. The idea behind lifted inference is to exploit such symmetries, and compute the product and sum only once. From the algorithmic perspective, lifted variable elimination eliminates only one $A_i$ variable, then *exponentiates* the resulting factor (see formula below), and then sums out $T$. Mathematically, Expression 6 is computed as follows:

$$\sum_T \left( \sum_{A_1} (\phi_1(A_1, S)\phi_2(T, A_1)) \right)^N \tag{7}$$

The way in which lifted variable elimination manipulates the set of variables $\{A_1, \ldots, A_N\}$ is called *lifted multiplication* and *lifted summing-out* (a.k.a. lifted elimination). Note that the number of operations required is now constant in $N$. Assuming $N$ is already known, the main operation here is computing the $N$-th power, which is $O(\log N)$ (logarithmic in $N$ if exact arithmetic is used, constant for floating point arithmetic). Thus, lifted variable elimination runs in $O(\log N)$ time in this case.

## 2.4 Lifted Inference: Exploiting Symmetries within Factors

Now consider a second elimination order, where we first eliminate $T$ and then the $A_i$:

$$\tilde{Pr}(S) = \sum_{A_1,\ldots,A_N} \sum_T \prod_{i=1}^N \phi_1(A_i, S) \prod_{i=1}^N \phi_2(T, A_i) = \sum_{A_1,\ldots,A_N} \prod_{i=1}^N \phi_1(A_i, S) \left( \sum_T \prod_{i=1}^N \phi_2(T, A_i) \right) \tag{8}$$

With this order, regular variable elimination works as follows. The inner summation (elimination of $T$) first multiplies all factors $\phi_2(T, A_i)$ into a factor $\phi_3(T, A_1, \ldots, A_N)$, and then sums out $T$:

$$\sum_T \prod_{i=1}^N \phi_2(T, A_i) = \sum_T \phi_3(T, A_1, \ldots, A_N) = \phi_3'(A_1, \ldots, A_N)$$

Note that $\phi_3$ is a function of $N+1$ binary variables, so its tabular representation has $2^{N+1}$ entries, which makes the cost of this elimination $O(2^{N+1})$. Substituting the computed $\phi'_3$ into Equation (8) yields:

$$\tilde{Pr}(S) = \sum_{A_1,\dots,A_N} \left( \prod_{i=1}^{N} \phi_1(A_i, S) \right) \phi'_3(A_1, \dots, A_N)$$

Now we can multiply $\phi'_3(A_1, \dots, A_N)$ by $\phi_1(A_1, S)$ and sum out $A_1$, then multiply the result by $\phi_1(A_2, S)$ and sum out $A_2$, and so on, until we obtain a factor $\phi'_4(S)$:

$$\tilde{Pr}(S) = \phi'_4(S)$$

This again involves $N$ multiplications and summations with exponential complexity. In summary, variable elimination computes the result in $O(2^{N+1})$.

This elimination order also has symmetries that lifted inference can exploit. Let us examine $\phi_3(T, A_1, \dots, A_N)$, the product of factors $\phi_2(T, A_i)$. For each assignment $T = t$ and $(A_1, \dots, A_N) = (a_1, \dots a_N) \in \{true, false\}^N$:

$$\phi_3(t, a_1, \dots, a_N) = \phi_2(t, a_1) \dots \phi_2(t, a_N)$$

Note that, since each $a_i \in \{true, false\}$, the multiplicands on the right hand side can have only one of two values, $\phi_2(t, true)$ or $\phi_2(t, false)$. That is, for each $a_i = true$ there is a $\phi_2(t, true)$, and similarly for each $a_i = false$, a $\phi_2(t, false)$. This means that, with $\mathcal{A}_t = \{A_i | a_i = true\}$ and $\mathcal{A}_f = \{A_i | a_i = false\}$, we can rewrite the above expression as:

$$\phi_3(t, a_1, \dots, a_N) = \prod_{a_i \in \mathcal{A}_t} \phi_2(t, true) \prod_{a_i \in \mathcal{A}_f} \phi_2(t, false) = \phi_2(t, true)^{|\mathcal{A}_t|} \phi_2(t, false)^{|\mathcal{A}_f|}.$$

This shows that to evaluate $\phi_3(T, A_1, \dots, A_N)$ it suffices to know how many $A_i$ are true (call this number $n_t$) and false ($n_f$); we do not need to know the value of each individual $A_i$. We can therefore restate $\phi_3$ in terms of a new variable $\#[\mathcal{A}]$, called a *counting variable*, the value of which is the two-dimensional vector $(n_t, n_f)$. Generally, $\#[\mathcal{A}]$ can take any value $(x, y)$ with $x, y \in \mathbb{N}$ and $x + y = N$. We call such a value a *histogram*. It captures the distribution of values among $\mathcal{A} = \{A_1, \dots, A_N\}$. The reformulation of a factor in terms of a counting variable is called *counting conversion*. Rewriting $\phi_3(T, A_1, \dots, A_N)$ as $\phi^*_3(T, \#[\mathcal{A}])$, we have

$$\phi^*_3(t, (n_t, n_f)) = \phi_2(t, true)^{n_t} \phi_2(t, false)^{n_f}.$$

$\phi^*_3$ has $2(N+1)$ possible input combinations (two values for $t$ and $N+1$ values for $(n_t, n_f)$, since $n_t + n_f = N$). It can be tabulated in time $O(N)$, using the recursive formula $\phi^*_3(t, (n_t + 1, n_f - 1)) = \phi^*_3(t, (n_t, n_f)) \cdot \phi_1(t, true) / \phi_2(t, false)$. Note that VE's computation of $\phi_3$ was $O(2^N)$.

Because $\phi^*_3$ has only $2(N+1)$ possible input states, instead of $2^{N+1}$, we can now eliminate $T$ in $O(N)$:

$$\sum_T \prod_{i=1}^{N} \phi_2(T, A_i) = \sum_T \phi^*_3(T, \#[\mathcal{A}]) = \phi'_3(\#[\mathcal{A}])$$

Using this result, we continue with the elimination:

$$\tilde{Pr}(S) = \sum_{A_1,\dots,A_N} \prod_{i=1}^{N} \phi_1(A_i, S) \; \phi_3'(\#[\mathcal{A}])$$

Using counting conversion a second time, we can reformulate the result of $\prod_{i=1}^{N} \phi_1(A_i, S)$ as $\phi_4(\#[\mathcal{A}], S)$, which gives:

$$\tilde{Pr}(S) = \sum_{A_1,\dots,A_N} \phi_4(\#[\mathcal{A}], S) \; \phi_3'(\#[\mathcal{A}]) = \sum_{A_1,\dots,A_N} \phi_{43}(\#[\mathcal{A}], S) \qquad (9)$$

In itself, the final summation still enumerates all $2^N$ joint states of variables $\mathcal{A}$, computes the histogram $(n_t, n_f)$ and $\phi_{43}((n_t, n_f), S)$ for each state, and adds up all the $\phi_{43}$. But we can do better: all states that result in the same histogram $(n_t, n_f)$ have the same value for $\phi_{43}((n_t, n_f), S)$, and we know exactly how many such joint states there are, namely $\binom{N}{n_t} = \frac{N!}{n_t! n_f!}$. We will call this the *multiplicity* of the histogram $(n_t, n_f)$, denoted $\text{MUL}((n_t, n_f))$. Thus, we can compute $\phi_{43}((n_t, n_f), S)$ just once for each histogram $(n_t, n_f)$ and multiply it by its multiplicity:

$$\sum_{A_1,\dots,A_N} \phi_{43}(\#[\mathcal{A}], S) = \sum_{\#[\mathcal{A}]} \text{MUL}(\#[\mathcal{A}]) \cdot \phi_{43}(\#[\mathcal{A}], S)$$

This way we enumerate over $N+1$ possible values of $\#[\mathcal{A}]$ instead of $2^N$ possible states of $\mathcal{A}$. To summarize, we can reformulate Equation (9) as

$$\tilde{Pr}(S) = \sum_{A_1,\dots,A_N} \phi_{43}(\#[\mathcal{A}], S) = \sum_{\#[\mathcal{A}]} \text{MUL}(\#[\mathcal{A}]) \cdot \phi_{43}(\#[\mathcal{A}], S) = \phi_5(S)$$

which shows that $\#[\mathcal{A}]$ can be eliminated with $O(N)$ operations.

The whole computation of $\tilde{Pr}(S)$ thus has complexity $O(N)$, instead of $O(2^N)$ for VE with this elimination order. This reduction in complexity is possible due to symmetries in the model that allow us to treat all variables $\mathcal{A}$ as one unit $\#[\mathcal{A}]$.

## 2.5 Capturing the Symmetries

It is clear that lifting can yield important speedups, if certain symmetries among factors or among the inputs of a single factor are present. To exploit these, it is essential that one can indicate which variables are interchangeable and hence induce these symmetries.

In our workshop example, assume, for instance, that not every person has the same preferences with respect to topics, but there are two types of people, and different potentials ($\phi_{2a}$ and $\phi_{2b}$) are associated with each type of person. It is clear that instead of Formula 7,

$$\sum_{T} \left( \sum_{A_1} \phi_1(A_1, S) \phi_2(T, A_1) \right)^N,$$

we then need to compute

$$\sum_T \left(\sum_{A_k} \phi_1(A_k, S)\phi_{2a}(T, A_k)\right)^{N_a} \left(\sum_{A_l} \phi_1(A_l, S)\phi_{2b}(T, A_l)\right)^{N_b}$$

where $A_k$ and $A_l$ are random members from the first and second group, and $N_a$ and $N_b$ the cardinality of these groups. In order to do this, we need to be able to state for which $A_i$ $\phi_{2a}$ is relevant, and for which $\phi_{2b}$ is. (For this particular computation, it actually suffices to know the size of each group, but that is not true in general; for instance, to compute the marginal distribution of $A_5$, we need to know which group $A_5$ is in.)

Our main contribution is related to this particular point. At the time of writing, the C-FOVE system (Milch et al., 2008) is considered the state of the art in lifted variable elimination. By introducing counting variables, it can capture within-factor symmetries better than its predecessor, FOVE. However, as it turns out, C-FOVE is less good at capturing symmetries among multiple factors, compared to FOVE. This is because groups of variables or factors are defined by means of constraints, and C-FOVE uses a constraint language that is more limited than FOVE's; essentially, it only allows for conjunctive constraints.

There are two reasons why it is important to be able to group variables with as much flexibility as possible. First, it gives more flexibility to the user who has to specify the graphical model itself. Second, during inference, it may become necessary to "split up" groups into subgroups.

We cannot go into detail about the constraint based representation at this point (we will do that later), but basically, during lifted inference, one may have a set of interchangeable variables that could in principle be treated as one group, but are not because the system cannot represent this group. It then needs to partition the group into smaller groups, possibly up to the level of individuals. For instance, assume the groups in our above example are $\{A_1, A_2, A_5, A_6, A_7\}$ and $\{A_3, A_4, A_8\}$. Further assume that the constraint language is such that sets of variables $A_i$ are defined using constraints of the form $\{A_i | l \leq i \leq u\}$. Neither group can be represented using one single constraint. For instance, the first group consists of the union of $\{A_i | 1 \leq i \leq 2\}$ and $\{A_i | 5 \leq i \leq 7\}$. Using this constraint language, we get four groups of size 2, 3, 2 and 1 instead of two groups of size 5 and 3. As a result, the computation actually performed will contain four exponentiated factors instead of two:

$$\sum_T \quad \left(\sum_{A_1} \phi_1(A_1, S)\phi_{2a}(T, A_1)\right)^2 \left(\sum_{A_5} \phi_1(A_5, S)\phi_{2a}(T, A_5)\right)^3$$
$$\left(\sum_{A_3} \phi_1(A_3, S)\phi_{2b}(T, A_3)\right)^2 \left(\sum_{A_8} \phi_1(A_8, S)\phi_{2b}(T, A_8)\right)^1$$

Generally, during lifted inference, groups may be split repeatedly. Unnecessary splits can substantially hurt efficiency, as each one causes a duplication of work. Since the duplicated work may include further splitting, the overall effect can be exponential in the number of consecutive splits.

Ideally, the constraint language should have the property that for each group of variables, there exists a constraint that represents exactly that group of variables. In that case, it is never necessary to split a group into subgroups just because the group cannot be represented. We call such a language "extensionally complete". The main contribution of this article is that it shows how to perform lifted variable elimination with an extensionally complete constraint language. To this aim, first, a lifted variable elimination algorithm is defined in a way that is independent of the constraint representation mechanism, by defining its operators in terms of relational algebra expressions. We call this algorithm GC-FOVE. To make GC-FOVE operational, some kind of constraint representation mechanism is of course needed. Any constraint language $\mathcal{L}$ can be plugged into GC-FOVE, as long as it is closed with respect to the relational algebra operators used by GC-FOVE. Second, we propose an extensionally complete constraint representation language that is based on trees. Such a language is necessarily closed with respect to the relational algebra operators, and therefore suitable for GC-FOVE. The resulting system, GC-FOVE[TREES], can perform inference at a higher level of granularity, and therefore more efficiently, than C-FOVE, which does not use an extensionally complete constraint language. The effect of this is in particular visible when evidence is given (which breaks symmetries and hence causes group splitting); in such cases, GC-FOVE achieves exponential speedups compared to C-FOVE.

This ends our informal introduction to lifted variable elimination and the main contribution this articles makes to it. In the following sections, we first introduce formal notation and terminology, then present our contributions in more detail.

## 3. Representation

Lifted inference exploits symmetries in a probabilistic model. Such symmetries often occur in models that have repeating structures, such as plates (Getoor & Taskar, 2007, Ch. 7), or, more generally, in probabilistic-logical models. Probabilistic-logical modeling languages (also called probabilistic-relational languages) combine the representational and inferential aspects of first-order logic with that of probability theory.

First-order logic languages refer to objects (possibly of various types) in some universe, and properties of, or relationships between, these objects. Formulas in these languages can express that some property holds for a particular object, or for an entire set of objects. For instance, the fact that all humans are mortal could be written as $\forall x : Human(x) \rightarrow Mortal(x)$. Probabilistic-logical models can, in a similar way, express probabilistic knowledge about all objects. For instance, they could state that for each human, there is a prior probability of 20% that he or she smokes: $\forall x : P(Smokes(x)|Human(x)) = 0.2$. It is this ability to make (probabilistic) statements about entire sets of objects that allows these languages to compactly express symmetries in a model. Many different languages exist for representing probabilistic-logical models (e.g., see Getoor & Taskar, 2007). We use a representation formalism based on undirected graphical models that is closely related to the one used in earlier work on lifted variable elimination (Poole, 2003; de Salvo Braz, 2007; Milch et al., 2008).

The concepts introduced in this section have also been introduced in earlier work (de Salvo Braz, 2007; Milch et al., 2008). Differences arise in terminology and notation as we emphasize the constraint part.

## 3.1 A Constraint-based Representation Formalism

An undirected model is a factorization of a joint distribution over a set of random variables (Kschischang et al., 2001). Given a set of random variables $\mathcal{X} = \{X_1, X_2, \ldots, X_n\}$, a factor consists of a potential function $\phi$ and an assignment of a random variable to each of $\phi$'s inputs. For instance, the factorization $f(X_1, X_2, X_3) = \phi(X_1, X_2)\phi(X_2, X_3)$ contains two different factors (even if their potential functions are the same).

Likewise, in our probabilistic-logical representation framework, a model is a set of factors. The random variables they operate on are properties of, and relationships between, objects in the universe. We now introduce some terminology to make this more concrete. We assume familiarity with set and relational algebra (union $\cup$, intersection $\cap$, difference $\setminus$, set partitioning, selection $\sigma_C$, projection $\pi_X$, attribute renaming $\rho$, join $\bowtie$); see, for instance, the work of Ramakrishnan and Gehrke (2003).

The term "variable" can be used in both the logical and probabilistic context. To avoid confusion, we use the term *logvar* to refer to logical variables, and *randvar* to refer to random variables. We write variable names in uppercase, and their values in lowercase. Sets or sequences of logvars are written in boldface, sets or sequences of randvars in calligraphic; their values are written in boldface lowercase.

The vocabulary of our representation includes a finite set of predicates and a finite set of constants. A *constant* represents an object in our universe. A *term* is either a constant or a logvar. A *predicate* $P$ has an arity $n$ and a finite range ($range(P)$); it is interpreted as a mapping from $n$-tuples of objects (constants) to the range. An *atom* is of the form $P(t_1, t_2, \ldots, t_n)$, where the $t_i$ are terms. A *ground atom* is an atom where all $t_i$ are constants. A ground atom represents a random variable; this implies that its interpretation, an element of $range(P)$, corresponds to the assignment of a value to the random variable. Hence, the range of a predicate corresponds to the range of the random variables it can represent, and is not limited to $\{true, false\}$ as in logic.

Logvars have a finite domain, which is a set of constants. The domain of a logvar $X$ is denoted $D(X)$. A *constraint* is a relation defined on a set of logvars, i.e., it is a pair $(\mathbf{X}, C_{\mathbf{X}})$, where $\mathbf{X} = (X_1, X_2, \ldots, X_n)$ is a tuple of logvars, and $C_{\mathbf{X}}$ is a subset of $D(\mathbf{X}) = \times_i D(X_i)$ (Dechter, 2003). Hence, $C_{\mathbf{X}}$ is a set, whose elements (tuples) indicate the allowed combinations of value assignments for the variables in $\mathbf{X}$. For ease of exposition, we identify a constraint with its relation $C_{\mathbf{X}}$, and write $C$ instead of $C_{\mathbf{X}}$ when $\mathbf{X}$ is apparent from the context. We assume an implicit ordering of values in $C_{\mathbf{X}}$'s tuples according to the order of logvars in $\mathbf{X}$. For instance with $\mathbf{X} = (X_1, X_2)$, the constraint $C_{\mathbf{X}} = \{(a, b), (c, d)\}$ indicates that there are two possibilities: either $X_1 = a$ and $X_2 = b$, or $X_1 = c$ and $X_2 = d$. A constraint that contains only one tuple is called *singleton*.

A constraint may be defined extensionally, by listing the tuples that satisfy it, or intensionally, by means of some logical condition, expressed in a *constraint language*. We call a constraint language $\mathcal{L}$ *extensionally complete* if it can express any relation over logvars $\mathbf{X}$, i.e., for any subset of $D(\mathbf{X})$, there is a constraint $C_{\mathbf{X}} \in \mathcal{L}$ whose extension is exactly that subset.

A *constrained atom* is of the form $P(\mathbf{X})|C$, where $P(\mathbf{X})$ is an atom and $C$ is a constraint on $\mathbf{X}$. A constrained atom $P(\mathbf{X})|C$ represents a set of ground atoms $\{P(\mathbf{x})|\mathbf{x} \in C\}$, and hence a set of randvars. For consistency with the literature, we call such a constrained atom a *parametrized randvar* (PRV), and use calligraphic notation to denote it. Given a PRV $\mathcal{V}$, we use $RV(\mathcal{V})$ to denote the set of randvars it represents; we also say these randvars are *covered* by $\mathcal{V}$.

A *valuation* of a randvar (set of randvars) is an assignment of a value to the randvar (an assignment of values to all randvars in the set).

**Example 1.** The PRV $\mathcal{V} = Smokes(X)|C$, with $C = \{x_1, \ldots, x_n\}$, represents $n$ randvars $\{Smokes(x_1), \ldots Smokes(x_n)\}$.

A *factor* $f = \phi_f(\mathcal{A}_f)$ consists of a sequence of randvars $\mathcal{A}_f = (A_1, \ldots, A_n)$ and a *potential* function $\phi_f : \times_{i=1}^{n} range(A_i) \rightarrow \mathbb{R}^+$. The product of two factors, $f_1 \otimes f_2$, is defined as follows. Factor $f = \phi(\mathcal{A})$ is the product of $f_1 = \phi_1(\mathcal{A}_1)$ and $f_2 = \phi_2(\mathcal{A}_2)$ if and only if $\mathcal{A} = \mathcal{A}_1 \cup \mathcal{A}_2$ and for all $\mathbf{a} \in D(\mathcal{A})$: $\phi(\mathbf{a}) = \phi_1(\mathbf{a}_1)\phi_2(\mathbf{a}_2)$ with $\pi_{\mathcal{A}_i}(\mathbf{a}) = \mathbf{a}_i$ for $i = 1, 2$. That is, $\mathbf{a}$ assigns to each randvar in $\mathcal{A}_i$ the same value as $\mathbf{a}_i$. We use $\prod$ to denote multiplication of multiple factors. Multiplying a factor by a scalar $c$ means replacing its potential $\phi$ by $\phi' : x \mapsto c \cdot \phi(x)$.

An *undirected model* is a set of factors $F$. It represents a probability distribution $\mathcal{P}_F$ on randvars $\mathcal{A} = \bigcup_{f \in F} \mathcal{A}_f$ as follows: $\mathcal{P}_F(\mathcal{A}) = \frac{1}{Z} \prod_{f \in F} \phi_f(\mathcal{A}_f)$, with $Z$ a normalization constant such that $\sum_{\mathbf{a} \in range(\mathcal{A})} \mathcal{P}_F(\mathbf{a}) = 1$.

A *parametric factor* or *parfactor* has the form $\phi(\mathcal{A})|C$, with $\mathcal{A} = \{A_i\}_{i=1}^{n}$ a sequence of atoms, $\phi$ a potential function on $\mathcal{A}$, and $C$ a constraint on the logvars appearing in $\mathcal{A}$.[1] The set of logvars occurring in $\mathcal{A}$ is denoted $logvar(\mathcal{A})$; the set of logvars in $C$ is denoted $logvar(C)$. A factor $\phi(\mathcal{A}')$ is a *grounding* of a parfactor $\phi(\mathcal{A})$ if $\mathcal{A}'$ can be obtained by instantiating $\mathbf{X} = logvar(\mathcal{A})$ with some $\mathbf{x} \in C$. The set of groundings of a parfactor $g$ is denoted $gr(g)$.

**Example 2.** Parfactor $g_1 = \phi_1(Smokes(X))|X \in \{x_1, \ldots, x_n\}$ represents the set of factors $gr(g_1) = \{\phi_1(Smokes(x_1)), \ldots, \phi_1(Smokes(x_n))\}$.

A set of parfactors $G$ is a compact way of defining a set of factors $F = \{f | f \in gr(g) \wedge g \in G\}$ and the corresponding probability distribution $\mathcal{P}_G(\mathcal{A}) = \frac{1}{Z} \prod_{f \in F} \phi_f(\mathcal{A}_f)$.

## 3.2 Counting Formulas

Milch et al. (2008) introduced the idea of counting formulas and (parametrized) counting randvars.

A *counting formula* is a syntactic construct of the form $\#_{X_i \in C}[P(\mathbf{X})]$, where $X_i \in \mathbf{X}$ is called the *counted logvar*.

A *grounded counting formula* is a counting formula in which all arguments of the atom $P(\mathbf{X})$, except for the counted logvar, are constants. It defines a *counting randvar* (CRV), the meaning of which is as follows. First, we define the set of randvars it *covers* as $RV(\#_{X \in C}[P(\mathbf{X})]) = RV(P(\mathbf{X})|X \in C)$. The value of the CRV is determined by the values of the randvars it covers. More specifically, it is a *histogram* that indicates, given a valuation of $RV(P(\mathbf{X})|X \in C)$, how many different values of $X$ occur for each $r \in range(P)$. Thus, its value is of the form $\{(r_1, n_1), (r_2, n_2), \ldots, (r_k, n_k)\}$, with $r_i \in range(P)$ and $n_i$ the corresponding count. Given a histogram $h$, we will also write $h(v)$ for the count of $v$ in $h$. Note that the range of a CRV, i.e., the set of all possible histograms it can take as a value, is determined by $k = |range(P)|$ and $|C|$.

---

1. We use the definition of Kisynski and Poole (2009a) for parfactors, as it allows us to simplify the notation.

**Example 3.** $\#_{X \in \{x_1, x_2, x_3\}}[P(X, y, z)]$ is a grounded counting formula. It covers the randvars $P(x_1, y, z)$, $P(x_2, y, z)$ and $P(x_3, y, z)$. It defines a CRV, the value of which is determined by the values of these three randvars; if $P(x_1, y, z) = true$, $P(x_2, y, z) = false$ and $P(x_3, y, z) = true$, the CRV takes the value $\{(true, 2), (false, 1)\}$.

The concept of a CRV is somewhat complicated. A CRV behaves like a regular randvar in some ways, but not all. It is a construct that can occur as an argument of a factor, like regular randvars, but in that role it actually stands for a set of randvars, all of which are arguments of the factor. A factor of the form $\phi^*(\cdots, \#_{X \in C}[P(\mathbf{X})], \cdots)$ is equivalent to a factor of the form $\phi(\cdots, P(\mathbf{X}_1), P(\mathbf{X}_2), \ldots, P(\mathbf{X}_k), \cdots)$, with $P(\mathbf{X}_i)$ all the instantiations of $\mathbf{X}$ obtainable by instantiating $X$ with a value from $C$, and with $\phi$ returning for any valuation of the $P(\mathbf{X}_i)$ the value that $\phi^*$ returns for the corresponding histogram.

**Example 4.** The factor $\phi^*(\#_{X \in \{x_1, x_2, x_3\}}[P(X, y, z)])$ is equivalent to a factor $\phi(P(x_1, y, z), P(x_2, y, z), P(x_3, y, z))$. If $\phi^*(\{(true, 2), (false, 1)\}) = 0.3$, this implies that $\phi(false, true, true) = \phi(true, false, true) = \phi(true, true, false) = 0.3$.

As illustrated in Section 2.4, counting formulas are useful for capturing symmetries within a potential function. Recall the workshop example. Whether a person attends a workshop depends on its topic, and this dependence is the same for each person. We can represent this with a single parfactor $\phi(T, A(X)) | X \in \{x_1, \ldots, x_n\}$ that represents $n$ ground factors. Eliminating $T$ requires multiplying these $n$ factors into a single factor $\phi'(T, A(x_1), A(x_2), \ldots, A(x_n))$ before summing out $T$. The potential function $\phi'$ is high-dimensional, so a tabular representation for it would be very costly. However, it contains a certain symmetry: $\phi'$ depends only on how many times each possible value for $A(x_i)$ occurs, not on where exactly these occur. By representing the factor using a potential function $\phi^*$ that has only two arguments, $T$ and the CRV $\#_{X \in \{x_1, \ldots, x_n\}}[A(X)]$, it can be represented more concisely, and computed more efficiently. For instance, to sum out $A(X)$, we do not need to enumerate all possible $(2^n)$ value combinations of the $A(x_i)$ and sum the corresponding $\phi'(T, A(x_1), \ldots, A(x_n))$, we just need to enumerate all possible $(n + 1)$ values for the histogram of $\#_{X \in \{x_1, \ldots, x_n\}}[A(X)]$ and sum the corresponding $\phi^*(T, \#_{X \in \{x_1, \ldots, x_n\}}[A(X)])$, each multiplied by its multiplicity.

Note the complementarity between PRVs and CRVs. While the randvars covered by a PRV occur in different factors, the randvars covered by a CRV occur in one and the same factor. Thus, PRVs impose a symmetry among different factors, whereas CRVs impose a symmetry within a single factor.

A *parametrized counting randvar* (PCRV) is of the form $\#_X[P(\mathbf{X})] | C_{\mathbf{X}}$. In this notation we write the constraint on the counted logvar $X$ as part of the constraint $C_{\mathbf{X}}$ on all variables in $\mathbf{X}$. Similar to the way in which a PRV defines a set of randvars through its groundings, a PCRV defines a set of CRVs through its groundings of all variables in $\mathbf{X} \setminus \{X\}$.

**Example 5.** $\#_Y[Friend(X, Y)] | C$ represents a set of CRVs, one for each $x \in \pi_X(C)$, indicating the number of friends $x$ has. If $C = D(X) \times D(Y)$ with $D(X) = D(Y) =$

$\{ann, bob, carl\}$, we might for instance have $\#_Y[Friend(ann, Y)]||C = \{(true, 1), (false, 2)\}$ (Ann has one friend, and two people are not friends with her).

Some definitions from the previous section need to be extended slightly in order to accommodate PCRVs. First, because CRVs are not regular randvars, they are not included in the set of randvars covered by the PRCV; that is, $RV(\#_{X_i}[P(\mathbf{X})]||C) = RV(P(\mathbf{X})|C)$. Second, since a counting formula "binds" the counted logvar (it is no longer a parameter of the resulting PCRV), we define $logvar(\#_{X_i}[P(\mathbf{X})]) = \mathbf{X} \setminus \{X_i\}$. Thus, generally, $logvar(\mathcal{A})$ refers to all the logvars occurring in $\mathcal{A}$, *excluding* the counted logvars. Note that $logvar(C)$ remains unchanged: it refers to all logvars in $C$, whether they appear as counted or not.

We end this section with two definitions that will be useful later on.

**Definition 1 (Count function)**  *Given a constraint $C_{\mathbf{X}}$, for any $\mathbf{Y} \subseteq \mathbf{X}$ and $\mathbf{Z} \subseteq \mathbf{X} - \mathbf{Y}$, the function $\text{COUNT}_{\mathbf{Y}|\mathbf{Z}} : C_{\mathbf{X}} \to \mathbb{N}$ is defined as follows:*

$$\text{COUNT}_{\mathbf{Y}|\mathbf{Z}}(t) = |\pi_{\mathbf{Y}}(\sigma_{\mathbf{Z} = \pi_{\mathbf{Z}}(t)}(C_{\mathbf{X}}))|$$

*That is, for any tuple $t$, this function tells us how many values for $\mathbf{Y}$ co-occur with $t$'s value for $\mathbf{Z}$ in the constraint. We define $\text{COUNT}_{\mathbf{Y}|\mathbf{Z}}(t) = 1$ when $\mathbf{Y} = \emptyset$.*

**Definition 2 (Count-normalized constraint)**  *For any constraint $C_{\mathbf{X}}$, $\mathbf{Y} \subseteq \mathbf{X}$ and $\mathbf{Z} \subseteq \mathbf{X} - \mathbf{Y}$, $\mathbf{Y}$ is count-normalized w.r.t. $\mathbf{Z}$ in $C_{\mathbf{X}}$ if and only if*

$$\exists n \in \mathbb{N} : \forall t \in C_{\mathbf{X}} : \text{COUNT}_{\mathbf{Y}|\mathbf{Z}}(t) = n.$$

*When such an $n$ exists, we call it the conditional count of $\mathbf{Y}$ given $\mathbf{Z}$ in $C_{\mathbf{X}}$, and denote it $\text{COUNT}_{\mathbf{Y}|\mathbf{Z}}(C_{\mathbf{X}})$.*

**Example 6.** Let $\mathbf{X}$ be $\{P, C\}$ and let the constraint $C_{\mathbf{X}}$ be $(P, C) \in \{(ann, eric), (bob, eric), (carl, finn), (debbie, finn), (carl, gemma), (debbie, gemma)\}$. Suppose $C_{\mathbf{X}}$ indicates the parent relationship: Ann is a parent of Eric, etc. Then $\{P\}$ is count-normalized w.r.t. $\{C\}$ because all children (i.e., all instantiations of $C$ in $C_{\mathbf{X}}$: Eric, Finn and Gemma) have two parents according to $C_{\mathbf{X}}$, or formally, for all tuples $t \in C_{\mathbf{X}}$ it holds that $\text{COUNT}_{\{P\}|\{C\}}(t) = 2$. Conversely, $\{C\}$ is not count-normalized w.r.t. $\{P\}$ because not all parents have equally many children. For instance, $\text{COUNT}_{\{C\}|\{P\}}((ann, eric)) = 1$ (Ann has 1 child), but $\text{COUNT}_{\{C\}|\{P\}}((carl, finn)) = 2$ (Carl has 2 children).

## 4. The GC-FOVE Algorithm: Outline

We now turn to the problem of performing lifted inference on models specified using the above representation. The algorithm we introduce for this is called GC-FOVE (for Generalized C-FOVE). At a high level, it is similar to C-FOVE (Milch et al., 2008), the current state-of-the-art system in lifted variable elimination, but it differs in the definition and implementation of its operators.

Recall how standard variable elimination works. It eliminates randvars one by one, in a particular order called the elimination order. Elimination consist of multiplying all factors the randvar occurs into one factor, then summing out the randvar.

Similarly, GC-FOVE visits PRVs (as opposed to individual randvars) in a particular order. Ideally, it eliminates each PRV by multiplying the parfactors in which it occurs into one parfactor, then summing out the PRV, using the *lifted multiplication* and *lifted summing-out* operators. However, these operators are not always immediately applicable: it may be necessary to refine the involved parfactors and PRVs to make them so. This is done using other operators, which we call *enabling operators*.[2]

A high-level description of GC-FOVE is shown in Algorithm 1. Like C-FOVE, it makes use of a number of operators, and repeatedly selects and performs one of the possible operators on one or more parfactors. It uses the same greedy heuristic as C-FOVE for this selection, choosing the operation with the minimum cost, where the cost of each operation is defined as the total size (number of rows in tabular form) of all the potentials it creates.

The main difference between C-FOVE and GC-FOVE is in the operators used. Four of GC-FOVE's operators (MULTIPLY, SUM-OUT, COUNT-CONVERT and GROUND-LOGVAR) are a straightforward generalization of a similar operator in C-FOVE, the difference being that we provide definitions that work for any constraint representation language that is closed under relational algebra, instead of definitions that are specific for the constraint language used by C-FOVE. Three other operators (EXPAND, COUNT-NORMALIZE and SPLIT) also have counterparts in C-FOVE, but need to be redefined more substantially because they directly concern constraint manipulation. The lifted absorption operator (ABSORB) is completely new.

GC-FOVE in itself does not specify a particular constraint language. In practice, constraints have to be represented one way or another, so some constraint representation mechanism has to be plugged in. In this article, we propose a tree-based representation mechanism for constraints. Important advantages of this mechanism are that, on the one hand, any extensional set can be represented by these trees, and on the other hand, constraints can still be manipulated efficiently.

The generalization of the operators, the new absorption operator, and the tree-based constraint language are the main contributions of this paper. Together, they greatly improve the efficiency of inference, as will be clear from the experimental section. Before describing the operators in detail, we illustrate the importance of using an expressive constraint language.

## 4.1 Constraint Language

In C-FOVE, a constraint is a set of pairwise (in)equalities between a single logvar and a constant, or between two logvars. Thus, in a single parfactor, C-FOVE can represent, for instance, $Friend(X, Y)|X \neq ann$, but not $Friend(X, Y)|(X, Y) \in \{(ann, bob), (bob, carl)\}$. Table 1 provides some more examples of PRVs that C-FOVE can/cannot represent, and Figure 3 illustrates this visually. Basically, C-FOVE can only use conjunctive constraints, not disjunctive ones, and C-FOVE's operators are defined to operate directly on this representation. GC-FOVE, on the other hand, allows a constraint to be any relation on the logvars, and can therefore handle all these PRVs. Because it has no restrictions whatsoever regarding the constraints it can handle, it can maximally exploit opportunities for lifting.

---

2. Technically speaking, multiplication is also an enabling operator as summing-out can only be applied after multiplication.

---

GC-FOVE
**Inputs**:
$G$: a model
$Q$: the query randvar
**Algorithm**:
*while* $G$ contains other randvars than $Q$:
    *if* there is a PRV $\mathcal{V}$ that can be eliminated by lifted absorption
        $G \leftarrow$ apply operator ABSORB to eliminate $\mathcal{V}$ in $G$
    *else if* there is a PRV $\mathcal{V}$ that can be eliminated by lifted summing-out
        $G \leftarrow$ apply SUM-OUT to eliminate $\mathcal{V}$ in $G$
    *else* apply an enabling operator (one of MULTIPLY, COUNT-CONVERT, EXPAND,
        COUNT-NORMALIZE, SPLIT or GROUND-LOGVAR) on some parfactors in $G$
*end while*
*return $G$*

---

Algorithm 1: Outline of the GC-FOVE algorithm.

| PRV | C-FOVE | GC-FOVE |
|---|---|---|
| $Friend(X, Y)$ | yes | yes |
| $Friend(ann, Y)$ | yes | yes |
| $Friend(X, Y)|X \neq ann$ | yes | yes |
| $Friend(X, Y)|X \in \{ann, bob\}$ | yes* | yes |
| $Friend(X, Y)|(X, Y) \in \{(ann, bob),\ (bob, carl)\})$ | no | yes |

Table 1: Examples of parametrized random variables that can / cannot be represented using a single constraint by C-FOVE. Though the fourth constraint (yes*) is disjunctive, C-FOVE can represent it using a conjunction of inequality constraints. This is not the case for the fifth constraint. GC-FOVE can represent all constraints.

The expressiveness of the constraint representation language, and the way the constraints are handled by the operators, are crucial to the efficiency of lifted variable elimination. The reason is that variables continuously need to be re-grouped (i.e., constraints need to be rewritten) during inference. For instance, we can multiply $\phi_1(P(X))|\{x_1, x_2, x_3\}$ and $\phi_2(P(X))|\{x_1, x_2, x_3\}$ directly, resulting in a parfactor of the form $\phi_{12}(P(X))|\{x_1, x_2, x_3\}$, but we cannot multiply $\phi_1(P(X))|\{x_1, x_2, x_3\}$ and $\phi_2(P(X))|\{x_2, x_3, x_4, x_5\}$ into a single parfactor because their PRVs do not match. The solution is to split constraints and parfactors so that matching parfactors arise. In this particular case, a model with three parfactors arises: $\phi_1(P(x_1))$, $\phi_{12}(P(X))|\{x_2, x_3\}$ and $\phi_2(P(X))|\{x_4, x_5\}$. GC-FOVE's operations result in this model. C-FOVE, however, when splitting constraints, separates off one tuple at a time ("splitting based on substitution", Milch et al., 2008), which here results in four parfactors: $\phi_1(P(x_1))$; $\phi_{12}(P(X))|X \neq x_1, X \neq x_4, X \neq x_5$; $\phi_2(P(x_4))$; and $\phi_2(P(x_5))$ (assuming the domain of $X$ is $\{x_1, x_2, \ldots, x_5\}$). In this case, C-FOVE could in fact represent the separate factors $\phi_2(P(x_4))$ and $\phi_2(P(x_5))$ as one parfactor $\phi_2(P(X))|X \neq x_1, X \neq x_2, X \neq x_3$, but it does not do so (only the intersection of two constraints is kept on the lifted level), and in general, for non-unary predicates, this is not possible, as Table 1 shows. Because of its restricted constraint language, C-FOVE often has to create finer-grained partitions than necessary. GC-FOVE, because it uses an extensionally complete constraint language, does not suffer from this problem.

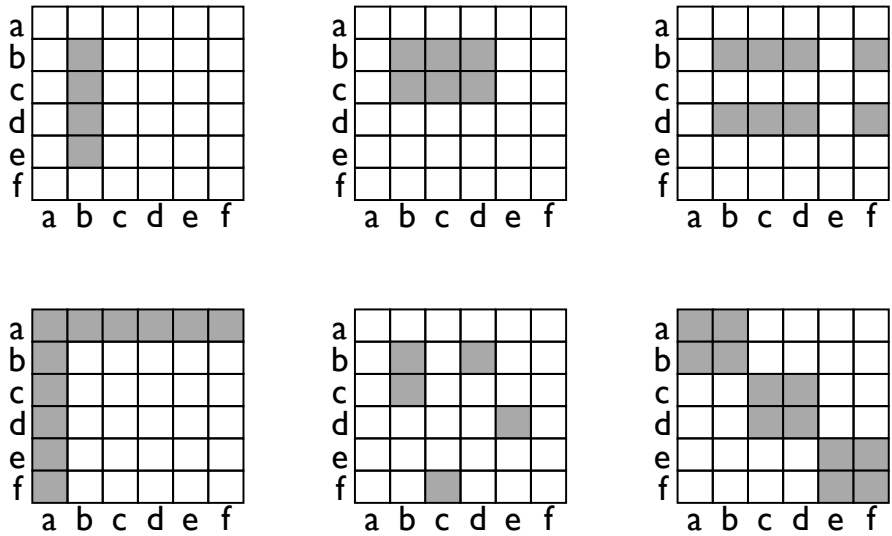

Figure 3: In each schema, the gray area indicates a PRV of the form $Friend(X, Y)|C_{XY}$ (with $a$ standing for *ann*, $b$ for *bob*, etc.) C-FOVE can only handle PRVs that can be defined by conjunctive constraints; this includes the top three schemas, but not the bottom ones. GC-FOVE can handle all PRVs.

## 4.2 Lifted Absorption

Absorption (van der Gaag, 1996) is an additional operator in VE that is known to increase efficiency. It consists of removing a random variable from a model when its valuation is known, and rewriting the model into an equivalent one that does not contain the variable. C-FOVE, like its predecessors, does not use absorption, and including it might in fact have detrimental effects due to breaking of symmetries. GC-FOVE's extensionally complete constraint language, however, not only makes it possible to use absorption more effectively, it even allows for lifting it.

## 4.3 Summary of Contributions

We are now at a point where we can summarize the contributions of this work more precisely.

1. We present the first description of lifted variable elimination that decouples the lifted inference algorithm from the constraint representation it uses. This is done by taking the C-FOVE algorithm and redefining its operators so that they become independent from the underlying constraint mechanism. This is achieved by defining the operators in terms of relational algebra operators. This redefinition generalizes the operators and clarifies on a higher level how they work.

2. We present a mechanism for representing constraints that is extensionally complete. It is closed under the relational algebra operators, and allows for executing them efficiently. In itself, this is a minor contribution, but it is necessary in order to obtain an operational system.

3. We present a new operator, called lifted absorption.

4. We experimentally demonstrate the practical impact of the above contributions.

5. We contribute the software itself.

Contributions 1 and 3 (our main contributions) are the subject of Section 5. Contribution 2 is detailed in Section 6, and Contribution 4 in Section 7. Contribution 5 is at `http://dtai.cs.kuleuven.be/ml/systems/gc-fove`.

## 5. GC-FOVE's Operators

This section provides detailed information on GC-FOVE's operators. These can conceptually be split into two categories: operators that manipulate potential functions, and operators that refine the model so that the first type of operators can be applied. We will start with three operators that belong to the first category: *lifted multiplication*, *lifted summing-out* and *counting conversion*. These can be seen as generalized versions of the corresponding C-FOVE operators; algorithmically, they are similar. Next, we discuss *splitting*, *shattering*, *expansion*, and *count normalization*. Because they operate specifically on the constraints, these differ more strongly from C-FOVE's operators. We will systematically compare them to the latter, showing each time that C-FOVE's constraint language and operators force it to create more fine-grained models than necessary, while GC-FOVE, because of its extensionally complete constraint language, can always avoid this: whatever the set of interchangeable randvars is, this set can be represented by one constraint. Finally, we discuss *lifted absorption*, which is completely new, and *grounding*, which is again similar to its C-FOVE counterpart.

In the following, $G$ refers to a model (i.e., a set of parfactors), and $G_1 \sim G_2$ means that models $G_1$ and $G_2$ define the same probability distribution.

### 5.1 Lifted Multiplication

The lifted multiplication operator multiplies whole parfactors at once, instead of separately multiplying the ground factors they cover (Poole, 2003; de Salvo Braz, 2007; Milch et al., 2008). Figure 4 illustrates this for two parfactors $g_1 = \phi_1(S(X))|C$ and $g_2 = \phi_2(S(X), A(X))|C$, where $C = (X \in \{x_1, \ldots, x_n\})$. Lifted multiplication is equivalent to $n$ multiplications on the ground level.

The above illustration is deceptively simple, for several reasons. First, the naming of the logvars suggests that logvar $X$ in $g_1$ corresponds to $X$ in $g_2$. In fact, $g_2$ could have multiple logvars, with different names. An *alignment* between the parfactors is necessary, showing how logvars in different parfactors correspond to each other (de Salvo Braz, 2007).

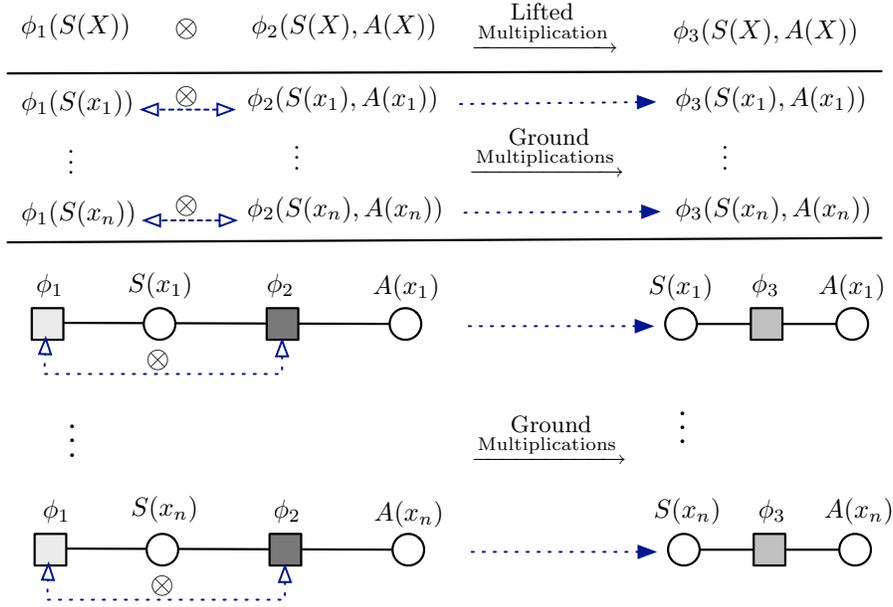

Figure 4: Lifted Multiplication with a 1:1 alignment between parfactors. The equivalent of the lifted operation (top), is shown at the level of ground factors (middle), and also in terms of factor graphs (bottom). $\otimes$ denotes (par)factor multiplication.

The alignment must constrain the aligned logvars to exactly the same values in $g_1$ and $g_2$ (otherwise, they cannot give identical PRVs in both parfactors). We formalize this as follows.

**Definition 3 (substitution)** A substitution $\theta = \{X_1 \to t_1, \ldots, X_n \to t_n\} = \{\mathbf{X} \to \mathbf{t}\}$ maps each logvar $X_i$ to a term $t_i$, which can be a constant or a logvar. When all $t_i$ are constants, $\theta$ is called a grounding *substitution, and when all are different logvars, a* renaming *substitution. Applying a substitution $\theta$ to an expression $\alpha$ means replacing each occurrence of $X_i$ in $\alpha$ with $t_i$; the result is denoted $\alpha\theta$.*

**Definition 4 (alignment)** An alignment $\theta$ between two parfactors $g = \phi(\mathcal{A})|C$ and $g' = \phi'(\mathcal{A}')|C'$ is a one-to-one substitution $\{\mathbf{X} \to \mathbf{X}'\}$, with $\mathbf{X} \subseteq logvar(\mathcal{A})$ and $\mathbf{X}' \subseteq logvar(\mathcal{A}')$, such that $\rho_\theta(\pi_{\mathbf{X}}(C)) = \pi_{\mathbf{X}'}(C')$ (with $\rho$ the attribute renaming operator).

An alignment tells the multiplication operator that two atoms in two different parfactors represent the same PRV, so it suffices to include it in the resulting parfactor only once. Including it twice is not wrong, but less efficient: some structure in the parfactor is then lost. For this reason, it is useful to look for "maximal" alignments which map as many PRVs to each other as possible.

**Example 7.** Consider $g_1 = \phi_1(S(X), F(X, Y))|C_{X,Y}$ and $g_2 = \phi_2(S(X'), F(X', Y'))|C_{X',Y'}$ with $C_{X,Y} = C_{X',Y'} = \{x_i\}_1^n \times \{y_j\}_1^m$. Using the maximal alignment $\{X \to X', Y \to$

$Y')\}$, we get the product parfactor $\phi_3(S(X), F(X, Y))|C_{X,Y}$. This alignment establishes a 1:1 association between each ground factor $\phi_1(S(x_i), F(x_i, y_j))$ and the corresponding $\phi_2(S(x_i), F(x_i, y_j))$. If, however, we multiply $g_1$ and $g_2$ with the alignment $\{X \rightarrow X'\}$, the result is a parfactor $\phi_3'(S(X), F(X, Y), F(X, Y'))|(X, Y, Y') \in \{x_i\}_1^n \times \{y_j\}_1^m \times \{y_k\}_1^m$, which for each $x_i$ unnecessarily multiplies each factor $\phi_1(S(x_i), F(x_i, y_j))$ with all factors $\phi_2(S(x_i), F(x_i, y_k)), k = 1, \ldots, m$. In other words, it unnecessarily creates a direct dependency between all pairs of randvars $F(x_i, y_j), F(x_i, y_k)$.

A second complication is that a single randvar may participate in multiple factors within a certain parfactor, and the number of such factors it appears in may differ across parfactors. Consider parfactors $g_1 = \phi_1(S(X))|X \in \{x_i\}_1^n$ and $g_2 = \phi_2(S(X), F(X, Y))|(X, Y) \in \{x_i\}_1^n \times \{y_i\}_1^m$. For each $x_i$, $\phi_1(S(x_i))$ shares randvar $S(x_i)$ with $m$ factors $\phi_2(S(x_i), F(x_i, y_j))$, $j = 1, \ldots, m$. Multiplication should result in a single parfactor $\phi_3(S(x_i), F(x_i, Y))|Y \in \{y_i\}_1^m$ that covers $m$ factors $\phi_3(S(x_i), F(x_i, y_j))$, and is equivalent to the product of one factor $\phi_1(S(x_i))$ and $m$ factors $\phi_2(S(x_i), F(x_i, y_j))$. This means we must find a $\phi_3$ such that $\forall v, w : \phi_3(v, w)^m = \phi_1(v) \prod_{i=1}^m \phi_2(v, w)$. This gives $\phi_3(v, w) = \phi_1(v)^{1/m} \phi_2(v, w)$. The exponentiation of $\phi_1$ to the power $1/m$ is called *scaling*. The result of this multiplication for a single $x_i$ is the same regardless of $x_i$, so finally, the product of the parfactors $g_1$ and $g_2$ will be the parfactor

$$\phi_3(S(X), F(X, Y)) = \phi_1(S(X))^{1/m} \cdot \phi_2(S(X), F(X, Y)) \, | \, (X, Y) \in \{x_i\}_1^n \times \{y_j\}_1^m.$$

Figure 5 illustrates this multiplication graphically.

An alignment between parfactors is called $1 : 1$ if all non-counted logvars in the parfactors are mapped to each other, and is called m:n otherwise. Multiplication based on an m:n alignment involves scaling, and requires that the non-aligned logvars be count-normalized (Definition 2, p. 406) with respect to the aligned logvars in the constraints (otherwise there is no single scaling exponent that is valid for the whole parfactor).

Operator 1 formally defines the lifted multiplication. Note that this definition does not assume any specific format for the constraints.

## 5.2 Lifted Summing-Out

Once a PRV occurs in only one parfactor, it can be summed out from that parfactor (Milch et al., 2008). We begin with an example of lifted summing-out, which will help motivate the formal definition of the operator.

**Example 8.** Consider parfactor $g = \phi(S(X), F(X, Y))|C$, in which $C = \{(x_i, y_{i,j}) : i \in \{1, \ldots, n\}, j \in \{1, \ldots, m\}\}$ (Figure 6). Note that $Y$ is count-normalized w.r.t. $X$ in $C$. Assume we want to sum out randvars $F(x_i, y_{i,j}) \in RV(F(X, Y)|C)$ on the ground level. Each randvar $F(x_i, y_{i,j})$ appears in exactly one ground factor $\phi(S(x_i), F(x_i, y_{i,j}))$ (see Figure 6 (middle)). We can therefore sum out each $F(x_i, y_{i,j})$ from its factor independently from the others, obtaining a factor $\phi'(S(x_i)) = \sum_{F(x_i, y_{i,j})} \phi(S(x_i), F(x_i, y_{i,j}))$. Since the $m$ ground factors $\phi(S(x_i), F(x_i, y_{i,j}))$ have the same potential $\phi$, summing out their second argument always results in the same potential $\phi'$, so we can compute $\phi'$ just once and, instead of storing $m$ copies of the resulting factor $\phi'(S(x_i))$, store a single factor $\phi''(S(x_i)) = \phi'(S(x_i))^m$. In the end, we obtain $n$ such factors, one for each $S(x_i)$, $i = 1, \ldots, n$. We can represent this result using a single parfactor $g' = \phi''(S(X))|C'$, with $C' = \{x_1, \ldots, x_n\} = \pi_X(C)$. Lifted summing-out directly computes $g'$ from $g$ in one operation. Note that to have a single exponent for all $\phi''$, $Y$ must be count-normalized w.r.t. $X$ in $C$.

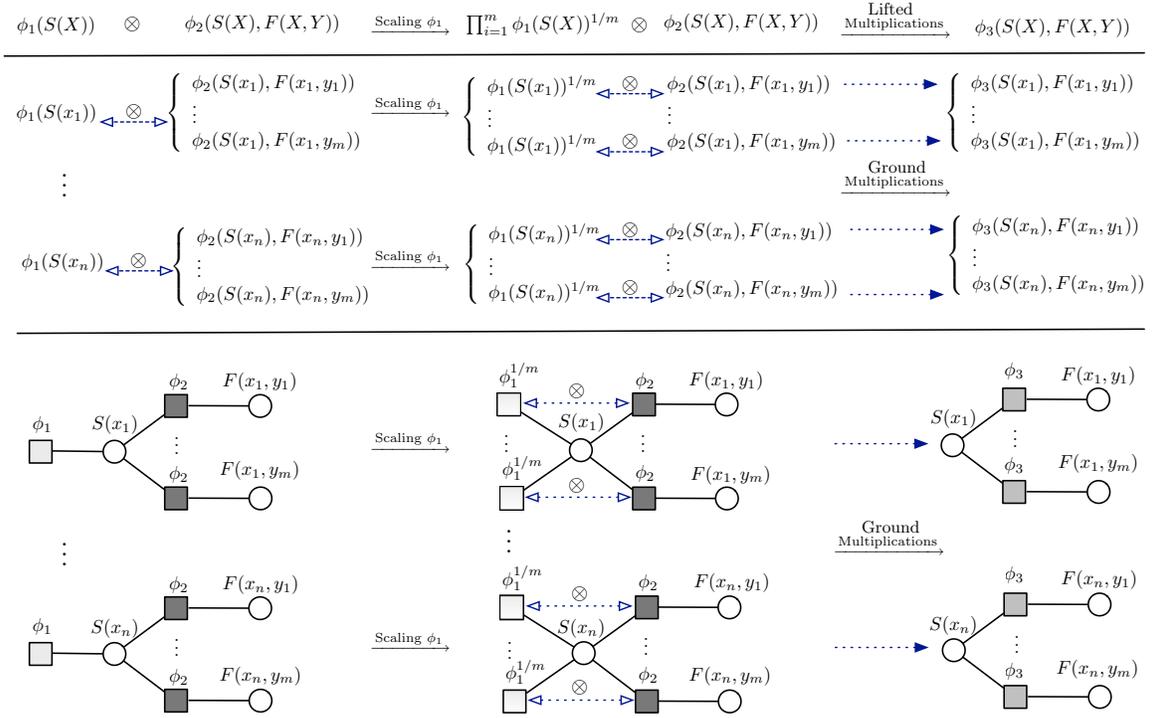

Figure 5: Lifted Multiplication with a m:n alignment between parfactors. The equivalent of the lifted operation (top), is shown at the level of ground factors (middle), and also in terms of factor graphs (bottom).

Like its C-FOVE counterpart, our lifted summing-out operator requires a one-to-one mapping between summed-out randvars and factors; that is, each summed-out randvar appears in exactly one factor, and all these factors are different. This is guaranteed when the eliminated atom contains all the logvars of the parfactor, since there is a different ground factor for each instantiation of the logvars. Further, lifted summing-out may result in identical factors on the ground level, which is exploited by computing one factor and exponentiating. This is the case when there is a logvar that occurs only in the eliminated atom, but not in the other atoms (such as $Y$ in $F(X, Y)$ in the above example).

As already illustrated in Section 2.4, counting randvars require special attention in lifted summing-out. A formula like $\phi(\#_X[P(X)])|X \in \{x_1, \ldots, x_k\}$ is really a shorthand for a factor $\phi(P(x_1), P(x_2), \ldots, P(x_k))$ whose value depends only on how many arguments take particular values. In principle, we need to sum out over all combinations of values of $P(X_i)$. We can replace this by summing out over all values of $\#_X[P(X)]$, on the condition that we take the multiplicities of the latter into account. The multiplicity of a histogram

---

**Operator** MULTIPLY
**Inputs:**
(1) $g_1 = \phi_1(\mathcal{A}_1)|C_1$: a parfactor in $G$
(2) $g_2 = \phi_2(\mathcal{A}_2)|C_2$: a parfactor in $G$
(3) $\theta = \{\mathbf{X}_1 \rightarrow \mathbf{X}_2\}$: an alignment between $g_1$ and $g_2$
**Preconditions:**
(1) for $i = 1, 2$: $\mathbf{Y}_i = logvar(\mathcal{A}_i) \setminus \mathbf{X}_i$ is count-normalized w.r.t. $\mathbf{X}_i$ in $C_i$
**Output**: $\phi(\mathcal{A})|C$, with
(1) $C = \rho_\theta(C_1) \bowtie C_2$.
(2) $\mathcal{A} = \mathcal{A}_1\theta \cup \mathcal{A}_2$, and
(3) for each valuation $\mathbf{a}$ of $\mathcal{A}$, with $\mathbf{a}_1 = \pi_{\mathcal{A}_1\theta}(\mathbf{a})$ and $\mathbf{a}_2 = \pi_{\mathcal{A}_2}(\mathbf{a})$ :
  $\phi(\mathbf{a}) = \phi_1^{1/r_2}(\mathbf{a}_1) \cdot \phi_2^{1/r_1}(\mathbf{a}_2)$, with $r_i = \text{COUNT}_{\mathbf{Y}_i|\mathbf{X}_i}(C_i)$
**Postcondition**: $G \sim G \setminus \{g_1, g_2\} \cup \{\text{MULTIPLY}(g_1, g_2, \theta)\}$

---

Operator 1: Lifted multiplication. The definition assumes, without loss of generality, that the logvars in the parfactors are standardized apart, i.e., the two parfactors do not share variable names (this can always be achieved by renaming logvars).

$h = \{(r_1, n_1), (r_2, n_2), \dots, (r_k, n_k)\}$ is a multinomial coefficient, defined as

$$\text{MUL}(h) = \frac{n!}{\prod_{i=1}^{k} n_i!}.$$

As multiplicities should only be taken into account for (P)CRVs, never for regular PRVs, we define for each PRV $A$ and for each value $v \in range(A)$: $\text{MUL}(A, v) = 1$ if $A$ is a regular PRV, and $\text{MUL}(A, v) = \text{MUL}(v)$ if $A$ is a PCRV. This MUL function is identical to Milch et al.'s (2008) NUM-ASSIGN.

With all this in mind, the formal definition of the lifted summing-out in Operator 2 is mostly self-explanatory. Precondition (1) ensures that all randvars in the summed-out P(C)RV occur exclusively in this parfactor. Precondition (2) ensures that each summed out randvar occurs in exactly one, separate, ground factor. Precondition (3) ensures that logvars occurring exclusively in the eliminated PRV are count-normalized with respect to the other logvars in that PRV, so that there is one unique exponent for exponentiation.

## 5.3 Counting Conversion

Counting randvars may be present in the original model, but they can also be introduced into parfactors by an operation called *counting conversion* (Milch et al., 2008) (see also Section 2.4). To see why this is useful, consider a parfactor $g = \phi(S(X), F(X,Y))|C$, with $C = \{x_i\}_{i=1}^n \times \{y_j\}_{j=1}^m$, and assume we want to eliminate $S(X)|C$. To do that, we first need to make sure each $S(x_i)$ occurs in only one factor. On the ground level, this can be achieved for a given $S(x_i)$ by multiplying all factors $\phi(S(x_i), F(x_i, y_j))$ in which it occurs. This results in a single factor $\phi'(S(x_i), F(x_i, y_1), \dots, F(x_i, y_m)) = \prod_j \phi(S(x_i), F(x_i, y_j))$ (see Figure 7). This is a high-dimensional factor, but because it equals a product of identical potentials $\phi$, its $F(x_i, y_j)$ arguments are mutually interchangeable: all that matters is how often values $v_1, v_2, \dots$ occur among them, not where they occur. This is exactly the kind of symmetry that CRVs aim to exploit. The factor $\phi'(S(x_i), F(x_i, y_1), \ldots, F(x_i, y_m))$ can therefore be replaced by a two-dimensional $\phi''(S(x_i), h)$ with $h$ a histogram that indicates how often each possible value in the range of $F(x_i, y_j)$ occurs. Thus, by introducing a CRV, we can define a two-dimensional $\phi''$ with that CRV as an argument, as opposed to the high-dimensional $\phi'$. As argued in Section 2.4, this reduces the size of the potential function, and hence computational complexity, exponentially.

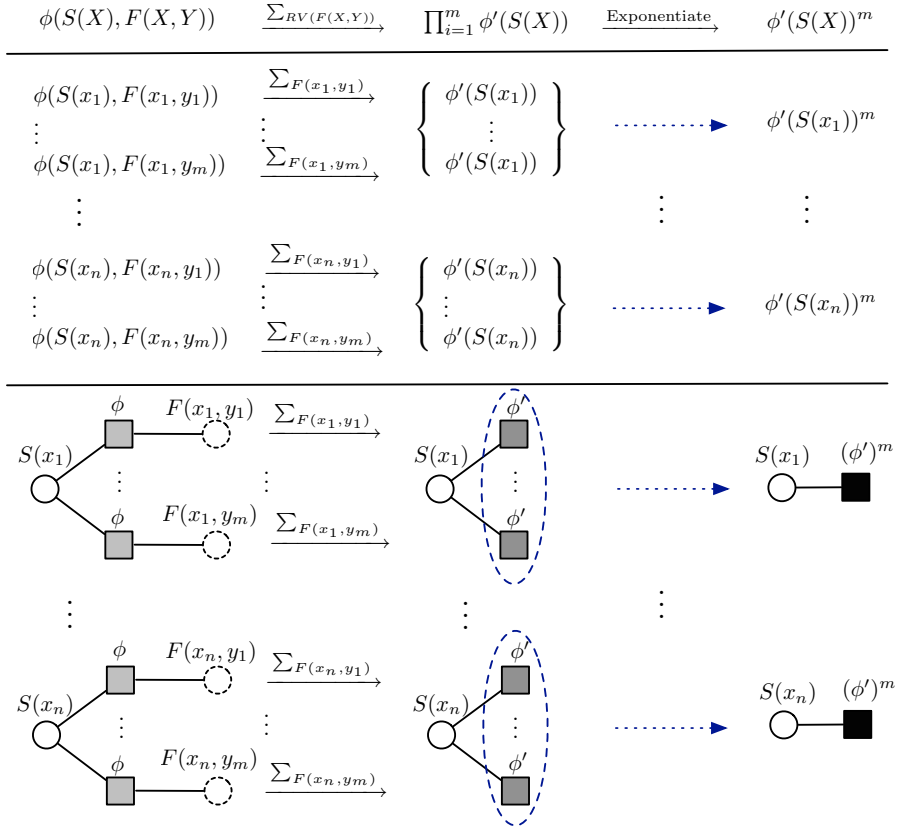

Figure 6: Lifted summing-out. The equivalent of the lifted operation (top), is shown at the level of ground factors (middle), and also in terms of factor graphs (bottom).

In many situations where lifted elimination cannot immediately be applied, counting conversion makes it applicable. The conditions of the SUM-OUT operator (Section 5.2) state that an atom $A_i$ can only be eliminated from a parfactor $g$ if $A_i$ has all the logvars in $g$. When an atom has fewer logvars than the parfactor, counting conversion modifies the parfactor by replacing another atom $A_j$ by a counting formula, which removes this counted logvar from $logvar(\mathcal{A})$. For instance, in the above example, $S(X)$ does not have the logvar $Y$ in $g = \phi(S(X), F(X, Y))|C$ and cannot be eliminated from the original parfactor $g$, but a counting conversion on $Y$ replaces $F(X, Y)$ with $\#_Y[F(X, Y)]$, allowing us to sum out $S(X)$ from the new parfactor $g' = \phi(S(X), \#_Y[F(X, Y)])|C$.

---

**Operator** SUM-OUT

**Inputs:**

(1) $g = \phi(\mathcal{A})|C$: a parfactor in $G$

(2) $A_i$: an atom in $\mathcal{A}$, to be summed out from $g_1$

**Preconditions**

(1) For all PRVs $\mathcal{V}$, other than $A_i|C$, in model $G$: $RV(\mathcal{V}) \cap RV(A_i|C) = \emptyset$

(2) $A_i$ contains all the logvars $X \in logvar(\mathcal{A})$ for which $\pi_X(C)$ is not singleton.

(3) $\mathbf{X}^{excl} = logvar(A_i) \setminus logvar(\mathcal{A} \setminus A_i)$ is count-normalized w.r.t.
   $\mathbf{X}^{com} = logvar(A_i) \cap logvar(\mathcal{A} \setminus A_i)$ in $C$

**Output**: $\phi'(\mathcal{A}')|C'$, such that

(1) $\mathcal{A}' = \mathcal{A} \setminus A_i$

(2) $C' = \pi_{X^{com}}(C)$

(3) for each assignment $\mathbf{a}' = (\ldots, a_{i-1}, a_{i+1}, \ldots)$ to $\mathcal{A}'$,
   $\phi'(\ldots, a_{i-1}, a_{i+1}, \ldots) = \sum_{a_i \in range(A_i)} \text{MUL}(A_i, a_i)\ \phi(\ldots, a_{i-1}, a_i, a_{i+1}, \ldots)^r$
   with $r = \text{COUNT}_{\mathbf{X}^{excl}|\mathbf{X}^{com}}(C)$

**Postcondition:** $\mathcal{P}_{G \setminus \{g\} \cup \{\text{SUM-OUT}(g, A_i)\}} = \Sigma_{RV(A_i|C)} \mathcal{P}_G$

---

Operator 2: The lifted summing-out operator.

Operator 3 formally defines counting conversion. It is mostly self-explanatory, apart from the preconditions. Precondition 1 makes sure that counting conversion, on the ground level, corresponds to multiplying factors that only differ in one randvar (i.e., are the same up to their instantiation of the counted logvar). Precondition 2 guarantees that the resulting histograms have the same range. Precondition 3 is more difficult to explain. It imposes a kind of independence between the logvar to be counted and already occurring counted logvars. Though not explicitly mentioned there, this precondition is also required for C-FOVE's counting operation; it implies that no inequality constraint should exist between $X$ and any counted logvar $X^{\#}$. A similar condition for FOVE's *counting elimination* is mentioned by de Salvo Braz (2007).

To see why precondition 3 is necessary, consider the parfactor $g = \phi(S(X), \#_Y[A(Y)])$ $|(X, Y) \in \{(x_1, y_2), (x_1, y_3), (x_2, y_1), (x_2, y_3), (x_3, y_1), (x_3, y_2)\}$, which does not satisfy it. This parfactor represents three factors of the form $\phi(S(x_i), \#_Y[A(Y)])|Y \in \{y_1, y_2, y_3\} \setminus \{y_i\}$, which contribute to the joint distribution with the product

$$\phi(S(x_1), \#_{Y \in \{y_2, y_3\}}[A(Y)]) \cdot \phi(S(x_2), \#_{Y \in \{y_1, y_3\}}[A(Y)]) \cdot \phi(S(x_3), \#_{Y \in \{y_1, y_2\}}[A(Y)]).$$

Counting conversion on logvar $X$ turns $g$ into a factor of the form

$$\phi'(\#_X[S(X)], \#_Y[A(Y)])$$

that should be equivalent. Note that $\phi'$ depends only on $\#_X[S(X)]$ and $\#_Y[A(Y)]$.

Now consider valuations $V_1$: $[S(x_1), S(x_2), S(x_3), A(y_1), A(y_2), A(y_3)] = [t, t, f, t, t, f]$ and $V_2$: $[S(x_1), S(x_2), S(x_3), A(y_1), A(y_2), A(y_3)] = [t, t, f, t, f, t]$. For both valuations, $\#_X[S(X)] = (2, 1)$ and $\#_Y[A(Y)] = (2, 1)$, so $\phi'(\#_X[S(X)], \#_Y[A(Y)])$ must return the same value under $V_1$ and $V_2$. The original parfactor, however, returns $\phi(S(t), (1, 1)) \cdot \phi(S(t), (1, 1)) \cdot \phi(S(f), (2, 0))$ under $V_1$, and $\phi(S(t), (1, 1)) \cdot \phi(S(t), (2, 0)) \cdot \phi(S(f), (1, 1))$ under $V_2$, which may be different. Since the original parfactor can distinguish valuations that no factor of the form $\phi'(\#_X[S(X)], \#_Y[A(Y)])$ can, counting conversion cannot be applied in this case.

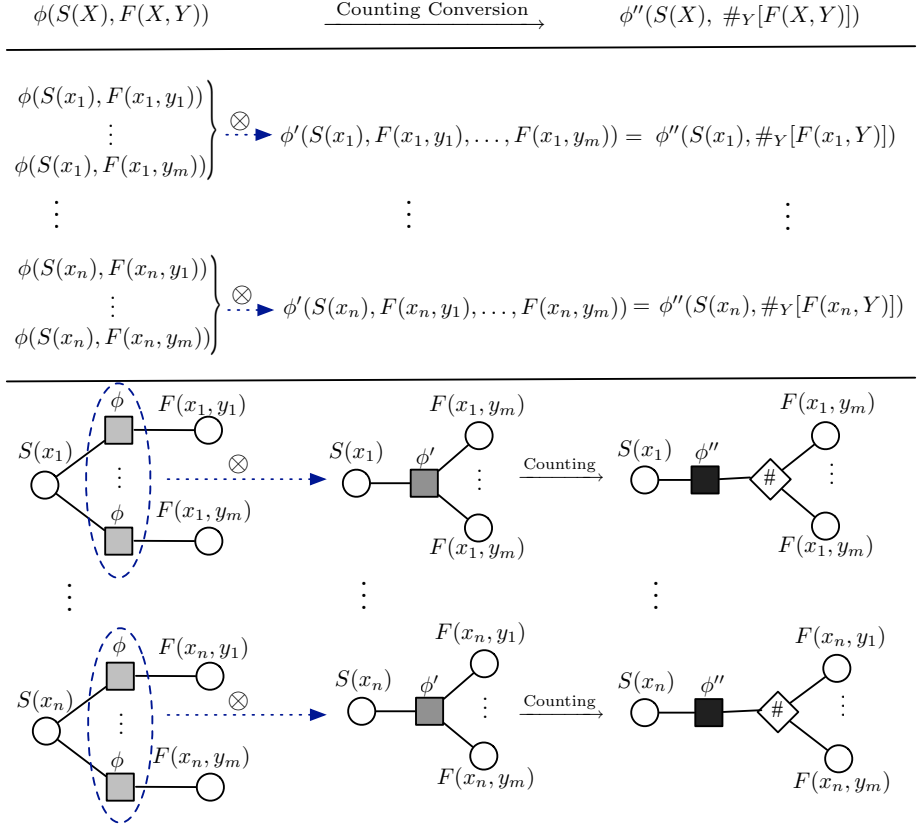

Figure 7: Counting conversion. The equivalent of the lifted operation (top), is shown at the level of ground factors (middle), and also in terms of factor graphs (bottom).

In contrast, consider $g' = \phi(S(X), \#_Y[A(Y)])|(X, Y) \in \{x_1, x_2, x_3\} \times \{y_1, y_2, y_3\}$, which is similar to $g$, except that its constraint satisfies precondition 3. All three factors represented by $g'$ differ only in their first argument, randvar $S(x_i)$; they have the same counting randvar $\#_Y[A(Y)]|Y \in \{y_1, y_2, y_3\}$ as their second argument (this was not the case for $g$). Their product, thus, can be represented by a parfactor $\phi'(\#_X[S(X)], \#_Y[A(Y)])|(X, Y) \in \{x_1, x_2, x_3\} \times \{y_1, y_2, y_3\}$, which is derived from $g'$ by a counting conversion.

## 5.4 Splitting and Shattering

When the preconditions for lifted multiplication, lifted summing-out and counting conversion are not fulfilled, it is necessary to reformulate the model in terms of parfactors that do fulfill them. For instance, if $g_1 = \phi_1(S(X))|X \in \{x_1, x_2, x_3\}$ and $g_2 = \phi_2(S(X))|X \in \{x_1, x_2, x_3, x_4, x_5\}$, we cannot multiply $g_1$ and $g_2$ directly without creating unwanted de-

---

**Operator** COUNT-CONVERT
**Inputs:**
(1) $g = \phi(\mathcal{A})|C$: a parfactor in $G$
(2) $X$: a logvar in $logvar(\mathcal{A})$
**Preconditions**
(1) there is exactly one atom $A_i \in \mathcal{A}$ with $X \in logvar(A_i)$
(2) $X$ is count-normalized w.r.t $logvar(\mathcal{A}) \setminus \{X\}$ in $C$
(3) for all counted logvars $X^{\#}$ in $g$: $\pi_{X,X^{\#}}(C) = \pi_X(C) \times \pi_{X^{\#}}(C)$
**Output:** $\phi'(\mathcal{A}')|C$, such that
(1) $\mathcal{A}' = \mathcal{A} \setminus \{A_i\} \cup \{A_i'\}$ with $A_i' = \#_X[A_i]$
(2) for each assignment $\mathbf{a}'$ to $\mathcal{A}'$ with $a_i' = h$:
  $\phi'(\dots, a_{i-1}, h, a_{i+1}, \dots) = \prod_{a_i \in range(A_i)} \phi(\dots, a_{i-1}, a_i, a_{i+1}, \dots)^{h(a_i)}$
  with $h(a_i)$ denoting the count of $a_i$ in histogram $h$
**Postcondition:** $G \sim G \setminus \{g\} \cup \{\text{COUNT-CONVERT}(g, X)\}$.

---

Operator 3: The counting conversion operator.

pendencies. However, we can replace $g_2$ with $g_{2a} = \phi_2(S(X))|X \in \{x_1, x_2, x_3\}$ and $g_{2b} = \phi_2(S(X))|X \in \{x_4, x_5\}$. The resulting model is equivalent, but in this new model, we can multiply $g_1$ with $g_{2a}$, resulting in $g_3 = \phi_3(S(X))|X \in \{x_1, x_2, x_3\}$.

The above is a simple case of *splitting* parfactors (Poole, 2003; de Salvo Braz, 2007; Milch et al., 2008). Basically, splitting two parfactors partitions each parfactor into a part that is shared with the other parfactor, and a part that is disjoint. The goal is to rewrite the P(C)RVs and parfactors into a *proper* form. Two P(C)RVs ($\mathcal{V}_1, \mathcal{V}_2$) are proper if $RV(\mathcal{V}_1)$ and $RV(\mathcal{V}_2)$ are either identical or disjoint; two parfactors are proper if all their P(C)RVs are proper. A pair of parfactors can be written into proper form by applying the following procedure, until all their P(C)RVs are proper. Choose a P(C)RV $\mathcal{V}_1$ from one parfactor, compare it to a P(C)RV $\mathcal{V}_2$ from the other, and rewrite the first parfactor such that $\mathcal{V}_1$ is split into two parts: one that is disjoint from $\mathcal{V}_2$ and one that is shared with $\mathcal{V}_2$. All the parfactors in the model can be made proper w.r.t. each other by repeatedly applying this rewrite until convergence. This is called *shattering* the model.

It is simpler to rewrite a PRV into the proper form than a PCRV. We describe the operator that handles PRVs, namely SPLIT, in this section and discuss the operator that handles PCRVs, namely EXPAND, in the following section. Before defining the SPLIT operator, we provide the following auxiliary definitions, which will also be used later on.

**Definition 5 (Splitting on overlap)** *Splitting a constraint $C_1$ on its $\mathbf{Y}$-overlap with $C_2$, denoted $C_1/_\mathbf{Y} C_2$, partitions $C_1$ into two subsets, containing all tuples for which the $\mathbf{Y}$ part occurs or does not occur, respectively, in $C_2$. $C_1/_\mathbf{Y} C_2 = \{\{t \in C_1 | \pi_\mathbf{Y}(t) \in \pi_\mathbf{Y}(C_2)\}, \{t \in C_1 | \pi_\mathbf{Y}(t) \notin \pi_\mathbf{Y}(C_2)\}\}$.*

**Definition 6 (Parfactor partitioning)** *Given a parfactor $g = \phi(\mathcal{A})|C$ and a partition $\mathbb{C} = \{C_i\}_{i=1}^n$ of $C$, PARTITION$(g, \mathbb{C}) = \{\phi(\mathcal{A})|C_i\}_{i=1}^n$.*

Operator 4 defines splitting of parfactors. Note that, in the operator definition, for simplicity, we assume that $A = A' = P(\mathbf{Y})$, which means that the logvars used in $A$ and

---

**Operator** SPLIT
**Inputs:**
(1) $g = \phi(\mathcal{A})|C$: a parfactor in $G$
(2) $A = P(\mathbf{Y})$: an atom in $\mathcal{A}$
(3) $A' = P(\mathbf{Y})|C'$ or $\#_Y[P(\mathbf{Y})]|C'$
**Output:** PARTITION$(g, \mathbb{C})$, with $\mathbb{C} = C/_{\mathbf{Y}}C' \setminus \{\emptyset\}$
**Postcondition** $G \sim G \setminus \{g\} \cup$ SPLIT$(g, A, A')$

---

Operator 4: The split operator.

$A'$ must be the same, in the same order. We can always rewrite the model such that any two PRVs with the same predicate are in this form. For this, we rewrite the parfactors as follows: (i) if the parfactors share logvars, we first standardize apart the logvars between two parfactors, (ii) *linearize* each atom in which some logvar occurs more than once, i.e., rewrite it such that it has a distinct logvar in each argument, and (iii) apply a renaming substitution on the logvars such that the concerned atoms have the same logvars. For instance, consider the two parfactors $g_1 = \phi_1(P(X, X))|X \in C_1$ and $g_2 = \phi_2(P(Y, Z))|(Y, Z) \in C_2$. The logvars of the two parfactors are already different, so there is no need for standardizing them apart. However, the atom $P(X, X)$ in $g_1$ is not linearized yet. To linearize it, we rewrite $g_1$ into the form $\phi_1(P(X, X'))|(X, X') \in C_1'$, where $C_1' = \{(x, x)|x \in C_1\}$. Finally, we rename the logvars $X$ and $X'$ to $Y$ and $Z$, respectively, to derive $\phi_1(P(Y, Z))|(Y, Z) \in C_1'$. This brings the atom $P(X, X)$ into the desired form $P(Y, Z)$.

For ease of exposition, we will not explicitly mention this linearization and renaming; whenever two PRVs from different parfactors are compared, any notation suggesting that they have the same logvars is to be interpreted as "have the same logvars after linearization and renaming".

When GC-FOVE wants to multiply two parfactors, it first checks for all pairs $A_1|C_1$, $A_2|C_2$ (one from each parfactor) whether they are proper. If a pair is found that is not proper, this means $A_1$ and $A_2$ are both of the form $P(\mathbf{Y})$, with different (but overlapping) instantiations for $\mathbf{Y}$ in $C_1$ and $C_2$. The pair is then split on $\mathbf{Y}$.

**Example 9.** Consider $g_1 = \phi_1(N(X, Y), R(X, Y, Z))|C_1$ with $C_1 = (X, Y, Z) \in \{x_i\}_{i=1}^{50} \times \{y_i\}_{i=1}^{50} \times \{z_i\}_{i=1}^{5}$, and $g_2 = \phi_2(N(X, Y))|C_2$ with $C_2 = (X, Y) \in \{x_{2i}\}_{i=1}^{25} \times \{y_i\}_{i=1}^{50}$. First, we compare the PRVs $N(X, Y)|C_1$ and $N(X, Y)|C_2$. These PRVs partially overlap, so splitting is necessary. To split the parfactors, we split $C_1$ and $C_2$ on their (X,Y)-overlap. This partitions $C_1$ into two sets: $C_1^{com} = \{x_{2i}\}_{i=1}^{25} \times \{y_i\}_{i=1}^{50} \times \{z_i\}_{i=1}^{5}$, and $C_1^{excl} = C_1 \setminus C_1^{com} = \{x_{2i-1}\}_{i=1}^{25} \times \{y_i\}_{i=1}^{50} \times \{z_i\}_{i=1}^{5}$. $C_2$ does not need to be split, as it has no tuples for which the (X,Y)-values do not occur in $C_1$. After splitting the constraints, we split the parfactors accordingly: $g_1$ is split into two parfactors $g_1^{com} = \phi(N(X, Y), R(X, Y, Z))|C_1^{com}$ and $g_1^{excl} = \phi(N(X, Y), R(X, Y, Z))|C_1^{excl}$, and parfactor $g_2$ remains unmodified.

Our splitting procedure splits any two PRVs into at most two partitions each. Similarly, the involved parfactors are split into at most two partitions each. This strongly contrasts with C-FOVE's approach to splitting. C-FOVE operates per logvar, and splits off each value in a separate partition (*splitting based on substitution*) (Poole, 2003; Milch et al., 2008). Thus, it may require many splits where GC-FOVE requires just one. In the above example, instead of $g_1^{excl} = \phi(N(X, Y), R(X, Y, Z))|C_1^{excl}$, C-FOVE ends up with 1250 parfactors $\phi(N(x_1, y_1), R(x_1, y_1, Z))|\{z_i\}_{i=1}^5$, $\phi(N(x_1, y_2), R(x_1, y_2, Z))|\{z_i\}_{i=1}^5$, ..., $\phi(N(x_3, y_1), R(x_3, y_1, Z))|\{z_i\}_{i=1}^5$, ..., $\phi(N(x_{49}, y_{50}), R(x_{49}, y_{50}, Z))|\{z_i\}_{i=1}^5$.

The reason why GC-FOVE can always split into at most two parfactors, yielding much coarser partitions than C-FOVE, is that it assumes an extensionally complete constraint language, whereas C-FOVE allows only pairwise (in)equalities, forcing it to split off each element separately.

## 5.5 Expansion of Counting Formulas

When handling parfactors with counting formulas, to rewrite a P(C)RV into the proper from, we employ the operation of *expansion* (Milch et al., 2008). When we split one group of randvars $RV(\mathcal{V})$ into a partition $\{RV(\mathcal{V}_i)\}_{i=1}^m$, any counting randvar $\gamma$ that counts the values of $RV(\mathcal{V})$ needs to be *expanded*, i.e., replaced by a group of counting randvars $\{\gamma_i\}_{i=1}^m$, where each $\gamma_i$ counts the values of randvars in $RV(\mathcal{V}_i)$. In parallel with this, the potential that originally had $\mathcal{V}$ as an argument must be replaced by a potential that has all the $\mathcal{V}_i$ as arguments; we call this *potential expansion*.

**Example 10.** Suppose we need to split $g_1 = \phi_1(\#_X[S(X)])|C_1$ and $g_2 = \phi_2(S(X))|C_2$, with $C_1 = \{x_1, \ldots, x_{100}\}$ and $C_2 = \{x_1, \ldots, x_{40}\}$. $C_1$ is split into $C_1^{com} = C_1 \cap C_2 = \{x_1, \ldots, x_{40}\}$ and $C_1^{excl} = C_1 \setminus C_2 = \{x_{41}, \ldots, x_{100}\}$. Consequently, the original group of randvars in parfactor $g_1$, namely $\{S(x_1), \ldots, S(x_{100})\}$, is partitioned into $\mathcal{V}_1^{com} = \{S(x_1), \ldots, S(x_{40})\}$ and $\mathcal{V}_1^{excl} = \{S(x_{41}), \ldots, S(x_{100})\}$. To preserve the semantics of the original counting formula, we now need two separate counting formulas, one for $\mathcal{V}_1^{com}$ and one for $\mathcal{V}_1^{excl}$, and we need to replace the original potential $\phi_1(\#_X[S(X)])$ by $\phi_1'(\#_{X_{com}}[S(X_{com})], \#_{X_{excl}}[S(X_{excl})])$, where $\phi_1'()$ depends only on the sum of the two new counting randvars $\#_{X_{com}}[S(X_{com})]$ and $\#_{X_{excl}}[S(X_{excl})]$. The end effect is that the parfactor $g_1$ is replaced by the new parfactor $\phi_1'(\#_{X_{com}}[S(X_{com})], \#_{X_{excl}}[S(X_{excl})])|C_1'$, where $C_1' = C_1^{com} \times C_1^{excl}$.

To explain expansion, we begin with the case of (non-parametrized) CRVs and then move to the general case of expansion for PCRVs.

### 5.5.1 EXPANSION OF CRVS

First consider the simplest possible type of CRV: $\#_X[P(X)]|C$. It counts for how many values of $X$ in $C$, $P(X)$ has a certain value. When $C$ is partitioned, $X$ must be counted within each subset of the partition.

In the following, we assume $C$ is partitioned into two non-empty subsets $C_1$ and $C_2$. If one of them is empty, the other equals $C$, which means the CRV can be kept as is and no expansion is needed.

In itself, splitting $\#_X[P(X)]|C$ into $\#_X[P(X)]|C_1$ and $\#_X[P(X)]|C_2$ is trivial, but a problem is that both of the resulting counting formulas will occur in one single parfactor, and a constraint is always associated with a parfactor, not with a particular argument of a parfactor. Thus, we need to transform $\phi(\#_X[P(X)])|C$ into a parfactor of the form $\phi'(\#_{X_1}[P(X_1)], \#_{X_2}[P(X_2)])|C'$, where the single constraint $C'$ expresses that $X_1$ can take only values in $C_1$, and $X_2$ only values in $C_2$. It is easily seen that $C' = \rho_{X \to X_1} C_1 \times \rho_{X \to X_2} C_2$ satisfies this condition. Further, to preserve the semantics, $\phi'$ should, for any count of $X_1$ and $X_2$, give the same result as $\phi$ with the corresponding count of $X$. The function

$$\phi'(h_1, h_2) = \phi(h_1 \oplus h_2),$$

with $\oplus$ denoting summation of histograms, has this property. Indeed, the histogram for $X_1$ (resp. $X_2$) in $C'$ is equal to that for $X$ in $C_1$ (resp. $C_2$), and since $\{C_1, C_2\}$ is a partition of $C$, the sum of these histograms equals the histogram for $X$ in $C$.

More generally, consider a non-parametrized CRV $\#_X[P(\mathbf{X})]|C$, with $X \in \mathbf{X}$ meaning that $\pi_{\mathbf{X}\setminus\{X\}}(C)$ is singleton. The constraint $C' = \pi_{\mathbf{X}\setminus\{X\}}(C) \times (\pi_{X_1}(\rho_{X \to X_1}C_1) \times \pi_{X_2}(\rho_{X \to X_2}C_2))$ joins this singleton with the Cartesian product of $\pi_X(C_1)$ and $\pi_X(C_2)$, and is equivalent to the constraint $\rho_{X \to X_1}(C_1) \bowtie \rho_{X \to X_2}(C_2)$. The result is again such that counting $X_1$ ($X_2$) in $C'$ is equivalent to counting $X$ in $C_1$ ($C_2$), while the constraint on all other variables remains unchanged. This shows that a parfactor $\phi(\mathcal{A}, \#_X[P(\mathbf{X})])|C$, for any partition $\{C_1, C_2\}$ of $C$ with $C_1$ and $C_2$ non-empty, can be rewritten in the form $\phi'(\mathcal{A}, \#_{X_1}[P(\mathbf{X})], \#_{X_2}[P(\mathbf{X})])|C'$, where $C' = \rho_{X \to X_1}(C_1) \bowtie \rho_{X \to X_2}(C_2)$.

Note that the ranges of the counting formulas in $\phi'$ (the $h_i$ arguments) depend on the cardinality of $C_1$ and $C_2$, which we will further denote as $n_1$ and $n_2$ respectively.

### 5.5.2 Expansion of PCRVs

Consider the case where $\pi_{\mathbf{X}\setminus\{X\}}(C)$ is not a singleton, i.e., we have a *parametrized* CRV $\mathcal{V}$ that represents a *group* of CRVs, each counting the values of a *subset* of $RV(\mathcal{V})$. Given a partitioning of the constraint $C$, we need to expand each underlying CRV and the corresponding potential. The constraint $C' = \rho_{X \to X_1}(C_1) \bowtie \rho_{X \to X_2}(C_2)$ remains correct (for non-empty $C_1, C_2$), even when $\pi_{\mathbf{X}\setminus\{X\}}(C)$ is no longer singleton: it associates the correct values of $X_1$ and $X_2$ with each tuple in $\pi_{\mathbf{X}\setminus\{X\}}(C)$. However, because the result of potential expansion depends on the size of the partitions, $n_1$ and $n_2$, only those CRVs that have the same $(n_1, n_2)$ result in identical potentials after expansion, and can be grouped in one parfactor. To account for this, PCRV expansion first splits the PCRV into groups of CRVs that have the same "joint count" $(n_1, n_2)$, then applies for each group the corresponding potential expansion.

To formalize this, we first provide the following auxiliary definitions.

**Definition 7 (Group-by)** *Given a constraint $C$ and a function $f : C \to R$, Group-By$(C, f) = C/\sim_f$, with $x \sim_f y \Leftrightarrow f(x) = f(y)$ and $/$ denoting set quotient. That is, Group-By$(C, f)$ partitions $C$ into subsets of elements that have the same result for $f$.*

**Definition 8 (Joint-count)** *Given a constraint $C$ over variables $\mathbf{X}$, partitioned into $\{C_1, C_2\}$, and a counted logvar $X \in \mathbf{X}$; then for any $t \in C$, with $L = \mathbf{X} \setminus \{X\}$ and $l = \pi_L(t)$,*

$$\text{joint-count}_{X,\{C_1, C_2\}}(t) = (|\pi_X(\sigma_{L=l}(C_1))|, |\pi_X(\sigma_{L=l}(C_2))|).$$

When a PCRV $\mathcal{V} = \#_{X_i}[P(\mathbf{X})]\,|C$ in a parfactor $g$ partially overlaps with another PRV $A'|C'$ in the model, expansion performs the following on $g$: (1) partition $C$ on its $\mathbf{X}$-overlap with $C'$, resulting in $C/\mathbf{X}C'$; (2) partition $C$ into $\mathbb{C} = \text{group-by}(C, \text{joint-count}_{X,C/\mathbf{X}C'})$ (this corresponds to a partition of $\mathcal{V}$ into CRVs that have the same number of rand-vars in each of the common and exclusive partitions in $C/\mathbf{X}C'$); (3) split $g$, based on

---

**Operator** EXPAND
**Inputs:**
(1) $g = \phi(\mathcal{A})|C$: a parfactor in $G$
(2) $A = \#_X[P(\mathbf{X})]$: a counting formula in $\mathcal{A}$
(3) $A' = P(\mathbf{X})|C'$ or $\#_Y[P(\mathbf{X})]|C'$
**Output:** $\{g_i = \phi_i'(\mathcal{A}_i')|C_i'\}_{i=1}^n$ where
(1) $C/_{\mathbf{X}}C' = \{C^{com}, C^{excl}\}$
(2) $\{C_1, \ldots, C_n\} = $ GROUP-BY$(C, $ JOINT-COUNT$_{X, C/_{\mathbf{X}}C'})$
(3) for all $i$ where $C_i \bowtie C^{com} = \emptyset$ or $C_i \bowtie C^{excl} = \emptyset$: $\phi_i' = \phi$, $\mathcal{A}_i' = \mathcal{A}$, $C_i' = C_i$
(4) for all other $i$:
$\quad$ (5) $C_i' = \pi_{logvar(\mathcal{A})}(C_i) \bowtie (\rho_{X \to X_{com}}(C^{com}) \bowtie \rho_{X \to X_{excl}}(C^{excl}))$
$\quad$ (6) $\mathcal{A}_i' = \mathcal{A} \setminus \{A\} \cup \{A\theta_{com}, A\theta_{excl}\}$ with $\theta_{com} = \{X \to X_{com}\}$, $\theta_{excl} = \{X \to X_{excl}\}$
$\quad$ (7) for each valuation $(\mathbf{l}, h_{com}, h_{excl})$ of $\mathcal{A}_i'$, $\phi_i'(\mathbf{l}, h_{com}, h_{excl}) = \phi(\mathbf{l}, h_{com} \oplus h_{excl})$
**Postcondition** $G \sim G \setminus \{g\} \cup $ EXPAND$(g, A, A')$

---

Operator 5: The expansion operator.

$\mathbb{C} = \{C_1, \ldots, C_n\}$, resulting in parfactors $g_1, \ldots, g_n$ that each require a distinct expanded potential; (4) in each $g_i$, replace potential $\phi$ with its expanded version. The formal definition of expansion is given in Operator 5.

**Example 11.** Suppose we need to split parfactors $g = \phi(\#_Y[F(X, Y)])|C$ and $g' = \phi'(F(X, Y))|C'$, with $C = \{ann, bob, carl\} \times \{dave, ed, fred, gina\}$ and $C' = \{ann, bob\} \times \{dave, ed\}$. Assume $F$ stands for friendship; $\#_Y[F(X, Y)]|C$ counts the number of friends and non-friends each $X$ has in $C$. The random variables covered by PCRV $\#_Y[F(X, Y)]\,|C$ partially overlap with those of $F(X, Y)\,|C'$. If we need to split $C$ on overlap with $C'$, yielding $C^{com}$ and $C^{excl}$, we need to replace the original PCRV with separate PCRVs for $C^{com}$ and $C^{excl}$. But PCRVs require count-normalization, and the fact that $Y$ is count-normalized w.r.t. $X$ in $C$ does not necessarily imply that the same holds in $C^{com}$ and $C^{excl}$. That is why, in addition to the split on overlap, we need an orthogonal partitioning of $C$ according to the joint counts. Within a subset $C_i$ of this partitioning, $Y$ will be count-normalized w.r.t. $X$ in $C_i^{com}$ and in $C_i^{excl}$.

We follow the four steps outlined above. Figure 8 illustrates these steps. First, we find the partition $C/_{X, Y}C' = \{C^{com}, C^{excl}\}$ with $C^{com} = \{ann, bob\} \times \{dave, ed\}$ and $C^{excl} = \{ann, bob\} \times \{fred, gina\} \cup \{carl\} \times \{dave, ed, fred, gina\}$. Inspecting the joint counts, we see that $C^{com}$ contains 2 possible friends for Ann or Bob (namely Dave and Ed), but 0 for Carl, whereas $C^{excl}$ contains 2 possible friends for Ann or Bob and 4 for Carl. Formally, JOINT-COUNT$_{Y, C/_{X, Y}C'}(t)$ equals (2,2) for $\pi_X(t) = ann$ or $\pi_X(t) = bob$, and equals (0,4) for $\pi_X(t) = carl$. So, within $C^{com}$ and $C^{excl}$, $Y$ is no longer count-normalized with respect to $X$. We therefore partition $C$ into subsets $\{C_1, C_2\} = $ GROUP-BY$(C, $ JOINT-COUNT$_{Y, C/_{X, Y}C'})$, which gives $C_1 = \{ann, bob\} \times \{dave, ed, fred, gina\}$ and $C_2 = \{carl\} \times \{dave, ed, fred, gina\}$. For each $C_i$, we can now construct a $C_i'$ that allows for counting the friends in $C_i^{com}$ and in $C_i^{excl}$ separately, using the series of joins discussed earlier. Where both $C_i^{com}$ and $C_i^{excl}$ are non-empty, the original PCRV $\#_Y[F(X, Y)]\,|C$ is

422

| $C$ | |
|---|---|
| ann | dave |
| ann | ed |
| ann | fred |
| ann | gina |
| bob | dave |
| bob | ed |
| bob | fred |
| bob | gina |
| carl | dave |
| carl | ed |
| carl | fred |
| carl | gina |

| $C'$ | |
|---|---|
| ann | dave |
| ann | ed |
| bob | dave |
| bob | ed |

| $C/_{X,Y} C'$ | |
|---|---|
| ann | dave |
| ann | ed |
| bob | dave |
| bob | ed |
| ann | fred |
| ann | gina |
| bob | fred |
| bob | gina |
| carl | dave |
| carl | ed |
| carl | fred |
| carl | gina |

GROUP-BY($C$, JOINT-COUNT$_{Y, C/_{X,Y} C'}$)

| | | |
|---|---|---|
| $C_1$ | ann | dave |
| | ann | ed |
| | ann | fred |
| | ann | gina |
| | bob | dave |
| | bob | ed |
| | bob | fred |
| | bob | gina |
| $C_2$ | carl | dave |
| | carl | ed |
| | carl | fred |
| | carl | gina |

| $C_1'$ | | |
|---|---|---|
| $X$ | $Y_{com}$ | $Y_{excl}$ |
| ann | dave | fred |
| ann | dave | gina |
| ann | ed | fred |
| ann | ed | gina |
| bob | dave | fred |
| bob | dave | gina |
| bob | ed | fred |
| bob | ed | gina |

| $C_2'$ | |
|---|---|
| $X$ | $Y$ |
| carl | dave |
| carl | ed |
| carl | fred |
| carl | gina |

Figure 8: Illustration of the PCRV expansion operator. (1) $Y$ is count-normalized w.r.t. $X$ in $C$ (with each $X$, four $Y$ values are associated). Splitting $C$ on overlap with $C'$ results in subsets in which $Y$ is no longer count-normalized w.r.t. $X$: the joint counts of $Y$ for both subsets are (2,2) for Ann and Bob, and (0,4) for Carl. To obtain count-normalized subsets, we need to partition $C$ into a subset $C_1$ for Ann and Bob, and $C_2$ for Carl; this is what the GROUP-BY construct does. For each of the subsets, a split on overlap with $C'$ will yield subsets in which $Y$ is count-normalized w.r.t. $X$. $C_1'$ is the result of joining the common and exclusive parts according to the join construct motivated earlier. $C_2'$ equals $C_2$ because $C_2$ has no overlap with $C'$ and hence need not be split.

replaced by two PCRVs per $C_i$, $\#_{Y_{com}}[F(X, Y_{com})] \,|\, C_i$ and $\#_{Y_{excl}}[F(X, Y_{excl})] \,|\, C_i$, and the new potential $\phi'$ is defined such that $\phi'(h_{com}, h_{excl}) = \phi(h_{com} \oplus h_{excl})$.

GC-FOVE's expansion improves over C-FOVE's in the following way. C-FOVE uses expansion based on substitution (Milch et al., 2008). For instance, in Example 10, C-FOVE splits off all the elements of $C^{excl}$ individually from $C$, adding each of these elements as a separate argument of the parfactor and the involved potential function. This yields a potential function $\phi'_1()$ with 61 arguments, namely the counting randvar $\#_{X_{com}}[S(X_{com})]$ and the 60 randvars $S(x_{41}), \ldots S(x_{100})$. This causes an extreme blow up in the size (number of entries) of the potential function, which does not happen using our approach. In general, C-FOVE's expansion yields a potential function of size $O(r^k \cdot (n-k)^r)$, with $n = |C_1|$, $k = |C_1^{excl}|$, and $r$ the cardinality of the range of the considered randvars (e.g., $r = |range(S(.))|$ in Example 10). In contrast, GC-FOVE's expansion yields a potential function of size $O(k^r \cdot (n-k)^r)$. In the likely scenario that $r \ll k$, this is exponentially smaller than C-FOVE's potential function. Given that this potential function will later be used for multiplication or summing-out, it is clear that GC-FOVE can yield large efficiency gains over C-FOVE.

## 5.6 Count Normalization

Lifted multiplication, summing-out and counting conversion all require certain variables to be count-normalized (recall Definition 2, p. 406). When this property does not hold, it can be achieved by *normalizing* the involved parfactor, which amounts to splitting the parfactor into parfactors for which the property does hold (Milch et al., 2008). Concretely, when $\mathbf{Y}$ is not count-normalized given $\mathbf{Z}$ in a constraint $C$, then $C$ is simply partitioned into $\mathbb{C} = \text{Group-By}(C, \text{Count}_{\mathbf{Y}|\mathbf{Z}})$, with $\text{Count}_{\mathbf{Y}|\mathbf{Z}}$ as defined in Definition 1; next, the parfactor is split according to $\mathbb{C}$. The formal definition of count normalization is shown in Operator 6.

---

**Operator** COUNT-NORMALIZE
**Inputs:**
(1) $g = \phi(\mathcal{A})|C$: a parfactor in $G$
(2) $\mathbf{Y}|\mathbf{Z}$: sets of logvars indicating the desired normalization property in $C$
**Preconditions**
(1) $\mathbf{Y} \subset logvar(\mathcal{A})$ and $\mathbf{Z} \subseteq logvar(\mathcal{A}) \setminus \mathbf{Y}$
**Output:** PARTITION($g$, GROUP-BY($C$, COUNT$_{\mathbf{Y}|\mathbf{Z}}$))
**Postconditions** $G \sim G \setminus \{g\} \cup \text{COUNT-NORMALIZE}(g, \mathbf{Y}|\mathbf{Z})$

---

Operator 6: The count-normalization operator.

**Example 12.** Consider the parfactor $g$ with $\mathcal{A} = (Prof(P), Supervises(P, S))$ and constraint $C = \{(p_1, s_1), (p_1, s_2), (p_2, s_2), (p_2, s_3), (p_3, s_5), (p_4, s_3), (p_4, s_4), (p_5, s_6)\}$. Lifted elimination of $Supervises(P, S)$ requires logvar $S$ (student) to be count-normalized with respect to logvar $P$ (professor). Intuitively, we need to partition the professors into groups such that all professors in the same group supervise the same number of students. In our example, $C$ needs to be partitioned into two, namely $C_1 = \sigma_{P \in \{p_3, p_5\}}(C) = \{(p_3, s_5), (p_5, s_6)\}$ (tuples involving professors with 1 student) and $C_2 = \sigma_{P \in \{p_1, p_2, p_4\}}(C) = \{(p_1, s_1), (p_1, s_2), (p_2, s_2), (p_2, s_3), (p_4, s_3), (p_4, s_4)\}$ (professors with 2 students). Next, the parfactor $g$ is split accordingly into two parfactors $g_1$ and $g_2$ with constraints $C_1$ and $C_2$. These parfactors are now ready for lifted elimination of $Supervises(P, S)$.

C-FOVE requires a stronger normalization property to hold. For every pair of logvars $X$ and $Y$ it requires either (1) $\pi_{X,Y}(C) = \pi_X(C) \times \pi_Y(C)$ or (2) $\pi_X(C) = \pi_Y(C)$ and $\pi_{X,Y}(C) =$ $(\pi_X(C) \times \pi_Y(C)) \setminus \{\langle x_i, x_i \rangle : x_i \in \pi_X(C)\}$. To enforce this, C-FOVE requires finer partitions than our approach does. In our example, C-FOVE requires $C$ to be split into 5 subsets $\{C_i\}_{i=1}^{5}$ with $C_i = \sigma_{P \in \{p_i\}}(C)$, i.e., one group per professor. The coarser partitioning used in our approach cannot be represented using C-FOVE's constraint language.

## 5.7 Absorption: Handling Evidence

When the value of a randvar is observed, this usually makes probabilistic inference more efficient: the randvar can be removed from the model, which may introduce extra independencies in the model. However, in lifted inference, there is also an adverse effect: observations can break symmetries among randvars. For this reason, it is important to handle observations in a manner that preserves as much symmetry as possible. In order to effectively handle observations in a lifted manner, we introduce the novel operator of *lifted absorption*.

In the ground setting, absorption works as follows (van der Gaag, 1996). Given a factor $\phi(\mathcal{A})$ and an observation $A_i = a_i$ with $A_i \in \mathcal{A}$, absorption replaces $\phi(\mathcal{A})$ with a factor $\phi'(\mathcal{A}')$, with $\mathcal{A}' = \mathcal{A} \setminus \{A_i\}$ and $\phi'(a_1, \ldots, a_{i-1}, a_{i+1}, \ldots, a_m) = \phi(a_1, \ldots, a_{i-1}, a_i, a_{i+1}, \ldots, a_m)$. This reduces the size of the factor and may introduce extra independencies in the model, which is always beneficial.

If $n$ randvars (built from the same predicate) have the same observed value, we can perform absorption on the lifted level by treating these $n$ randvars as one single group. Consider a parfactor $g = \phi(S(X), F(X, Y)) | (X, Y) \in \{(x_1, y_1), \ldots, (x_1, y_{50})\}$. Assume that evidence atoms $F(x_1, y_1)$ to $F(x_1, y_{10})$ all have the value true. This can be represented by adding an *evidence parfactor* $g_E$ to the model: $g_E = \phi_E(F(X, Y)) | (X, Y) \in \{x_1\} \times \{y_j\}_1^{10}$,[10] with $\phi_E(true) = 1$ (the observed value) and $\phi_E(false) = 0$. To absorb the evidence, $g$ needs to be split into two, namely $g_1$ with $C_1 = \{(x_1, y_1), \ldots, (x_1, y_{10})\}$ (the parfactor about which we have evidence) and $g_2$ with $C_2 = \{(x_1, y_{11}), \ldots, (x_1, y_{50})\}$ (no evidence). Then, we can absorb the evidence about $F$ into $g_1$. Performing absorption on the ground level would result in ten identical factors $\phi'(S(x_1))$ (the logvar $Y$ disappears in the absorption). Lifted absorption computes the same $\phi'$ once, and raises it to the tenth power. Generally, with $\mathbf{X}^{excl}$ the logvars that occur exclusively in the atom being absorbed, the exponent is the number of values $\mathbf{X}^{excl}$ can take, so $\mathbf{X}^{excl}$ must be count-normalized with respect to the other logvars. Further, all logvars in $\mathbf{X}^{excl}$ can be removed from the constraint $C$ as they disappear in the absorption.

For parfactors with counting formulas, essentially the same reasoning is used, but now the exponent is determined by the *non-counted* logvars occurring exclusively in the atom ($\mathbf{X}^{nce}$). These logvars, together with the counted logvar, can be removed from $C$. The value for the absorbed counting formula, to be filled in in $\phi$, is a histogram indicating how many times each possible value has been observed in the absorbed PRV. Since there is only one observed value in the evidence parfactor, this histogram maps that value to the number of randvars being absorbed, and other values to zero. Lifted absorption is formally defined in Operator 7. We provide a correctness proof for this operator in Appendix A, and analyze its complexity in Appendix B.

GC-FOVE handles evidence by absorption as follows. It first creates one evidence parfactor per observed value for each predicate. Next, it compares each evidence parfactor with

---

**Operator** ABSORB

**Inputs:**

(1) $g = \phi(\mathcal{A})|C$: a parfactor in $G$

(2) $A_i \in \mathcal{A}$ with $A_i = P(\mathbf{X})$ or $A_i = \#_{X_i}[P(\mathbf{X})]$

(3) $g_E = \phi_E(P(\mathbf{X}))|C_E$: an evidence parfactor

**Let** $\mathbf{X}^{excl} = \mathbf{X} \setminus logvar(\mathcal{A} \setminus A_i)$;

$\quad \mathbf{X}^{nce} = \mathbf{X}^{excl} \setminus \{X_i\}$ if $A_i = \#_{X_i}[P(\mathbf{X})]$, $\mathbf{X}^{excl}$ otherwise;

$\quad L' = logvar(\mathcal{A}) \setminus \mathbf{X}^{excl}$;

$\quad o$ = the observed value for $P(\mathbf{X})$ in $g_E$;

**Preconditions**

(1) $RV(A_i|C) \subseteq RV(A_i|C_E)$

(2) $\mathbf{X}^{nce}$ is count-normalized w.r.t. $L'$ in $C$.

**Output:** $g' = \phi'(\mathcal{A}')|C'$, with

(1) $\mathcal{A}' = \mathcal{A} \setminus \{A_i\}$

(2) $C' = \pi_{logvar(C) \setminus \mathbf{X}^{excl}}(C)$

(3) $\phi'(\dots, a_{i-1}, a_{i+1}, \dots) = \phi(\dots, a_{i-1}, e, a_{i+1}, \dots)^r$, with $r = \text{COUNT}_{\mathbf{X}^{nce}|L'}(C)$, and
with $e = o$ if $A_i = P(X)$
and $e$ a histogram with $e(o) = \text{COUNT}_{X_i|logvar(\mathcal{A})}(C)$, $e(.) = 0$ elsewhere, otherwise
(namely if $A_i = \#_{X_i}[P(\mathbf{X})]$)

**Postcondition**

$G \cup \{g_E\} = G \setminus \{g\} \cup \{g_E, \text{ABSORB}(g, A_i, g_E)\}$

---

Operator 7: Lifted absorption.

each PRV in the model, applying absorption when possible. Where necessary, parfactors in the model are split to allow for absorption. (It is never necessary to split evidence parfactors, see precondition 1.) When no more absorptions are possible with a given evidence parfactor, it is removed from the model: the evidence has been incorporated completely.

Like the SUM-OUT operator, the ABSORB operator has the effect of eliminating PRVs from the model. As the operator's definitions show, however, ABSORB requires weaker preconditions than SUM-OUT, which means that it can be applied in more situations. Also, the ABSORB operator easily lends itself to a *splitting as needed* constraint processing strategy (Kisynski & Poole, 2009a), which keeps the model at a much higher granularity, by requiring fewer splits on the parfactors compared to a preemptive shattering strategy. In the presence of observations, which is often the case in real-world problems, these effects can result in large computational savings.

Our approach to dealing with evidence differs from C-FOVE's in two important ways. First, C-FOVE introduces a separate evidence factor for each *ground* observation $A = a$. This causes extensive splitting: if there are $n$ randvars with the same observed value, there will be $n$ separate factors, and C-FOVE will perform (at least) $n$ eliminations on these randvars. In addition, the splitting may cause further splitting as C-FOVE continues, destroying even more opportunities for lifting. We show in Section 7 that this can make inference impossible with C-FOVE in the presence of evidence.

Second, C-FOVE does not use absorption; during inference, the evidence factors are used for multiplication and summing-out like any other factors. Absorption is advantageous

---

**Operator** GROUND-LOGVAR
**Inputs:**
(1) $g = \phi(\mathcal{A})|C$: a parfactor in $G$
(2) $X$: a logvar in $logvar(\mathcal{A})$
**Output:** PARTITION($g$, GROUP-BY($C, \pi_X$))
**Postcondition**
$G \sim G \setminus \{g\} \cup$ GROUND-LOGVAR($g, X$)

---

Operator 8: Grounding.

because it eliminates randvars from the model, so they no longer need to be summed out. As a result, in our approach, evidence reduces the number of summing-out and multiplication operations, while in C-FOVE it increases that number.

### 5.8 Grounding a Logvar

There is no guarantee that the enabling operators eventually result in PRVs and parfactors that allow for any of the lifted operators. To illustrate this, consider a model consisting of a single parfactor $\phi(R(X, Y), R(Y, Z), R(X, Z))|C$, which expresses a probabilistic variant of transitivity. Since there is only one factor, no multiplications are needed before starting to eliminate variables. Yet, because of the structure of the parfactor, no single PRV can be eliminated (the preconditions for lifted summing out and counting conversion are not fulfilled, and none of the other operators can change that).

In cases like this, when no other operators can be applied, lifted VE can always resort to a last operator: grounding a logvar $X$ in a parfactor $g$ (de Salvo Braz, 2007; Milch et al., 2008). Given a parfactor $g = \phi(\mathcal{A})|C$ and a logvar $X \in logvar(\mathcal{A})$ with $\pi_X(C) = \{x_1, \ldots, x_n\}$, grounding $X$ replaces $g$ with the set of parfactors $\{g_1, \ldots, g_n\}$ with $g_i = \phi(\mathcal{A})|\sigma_{X=x_i}(C)$. This is equivalent to splitting $g$ based on the partition GROUP-BY($C, \pi_X$), which yields the definition shown in Operator 8. Note that in each resulting parfactor $g_i$, logvar $X$ can only take on a single value $x_i$, so in practice $X$ can be replaced by the constant $x_i$ and removed from the set of logvars.

Grounding can significantly increase the granularity of the model and decrease the opportunities for performing lifted inference: in the extreme case where all logvars are grounded, inference is performed at the propositional level. It is therefore best used only as a last resort. In practice, (G)C-FOVE's heuristic for selecting operators, which relies on the size of the resulting factors, automatically has this effect.

Calling the GROUND-LOGVAR operator should not be confused with the event of obtaining a ground model. GROUND-LOGVAR grounds only one logvar, and does not necessarily result in a ground model. Conversely, one may arrive at a ground model without ever calling GROUND-LOGVAR, simply because the splitting continues up to the singleton level.

## 6. Representing and Manipulating the Constraints

We have shown that using an extensionally complete constraint language instead of allowing only pairwise (in)equalities can potentially yield large efficiency gains by allowing more opportunities for lifting. The question remains how we can represent these constraints.

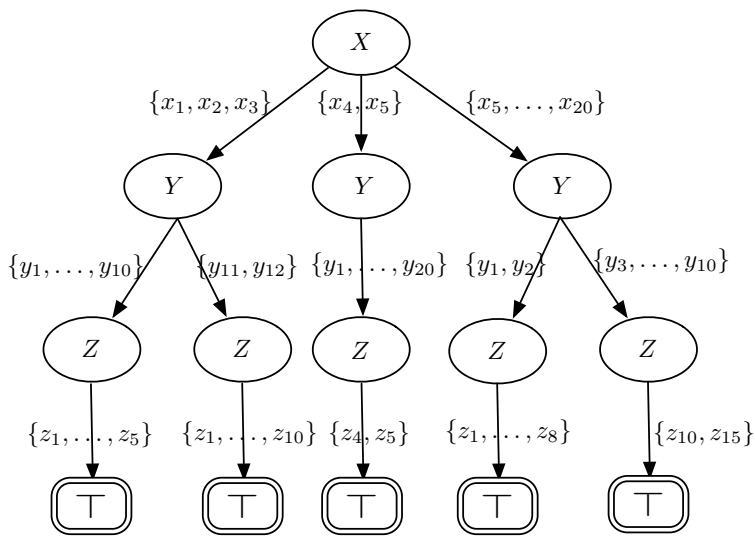

Figure 9: A constraint tree representing a constraint on logvars $X, Y, Z$.

In principle, we could represent them extensionally, as lists of tuples. This allows any constraint to be represented, but is inefficient when we have many logvars. Instead, we employ a *constraint tree*, as also used in First Order Bayes-Ball (Meert, Taghipour, & Blockeel, 2010). Hence, the lower-level operations on constraints (projection, splitting, counting) must be implemented in terms of constraint trees. Below, we briefly explain how this is done.

A constraint tree on logvars $\mathbf{X}$ is a tree in which each internal (non-leaf) node is labeled with a logvar $X \in \mathbf{X}$, each leaf is labeled with a terminal label $\top$, and each edge $e = (X_i, X_j)$ is labeled with a (sub-)domain $D(e) \subseteq D(X_i)$. See Figure 9 for an example. We use ordered trees, where all nodes in the same level of the tree are labeled with the same logvar, and each logvar occurs on only one level. Each path from the root to a leaf through edges $e_1, \ldots, e_{|\mathbf{X}|}$ represents the tuples in the Cartesian product $\times_i D(e_i)$. For example, in Figure 9, the left most path represents the tuples $\{x_1, x_2, x_3\} \times \{y_1, \ldots, y_{10}\} \times \{z_1, \ldots, z_5\}$. The constraint represented by the tree is the union of tuples represented by each root-to-leaf path.

Given a constraint (in terms of the set of tuples that satisfy it), we construct the corresponding tree in a bottom-up manner by merging compatible edges. Different logvar orders can result in trees of different sizes. A tree can be *re-ordered* by interchanging nodes in two adjacent levels of the tree and applying the possible merges at those levels. We employ re-ordering to simplify the various constraint handling operations. For *projection* of a constraint, we move the projected logvars to the top of the tree and discard the parts below these logvars. For *splitting*, we perform a pairwise comparison of the two involved constraint trees. First, we re-order each tree such that the logvars involved in the split are at the top of the trees. Then we process the trees top-down by comparing the edges leaving the root in the two trees and partition their domains based on their overlap. We recursively repeat this for their children until we reach the last logvar involved in the split. For *count normalization*, we also first apply this re-ordering. Then we partition the tree based on the number of tuples of counted logvars in each branch. For counting this number, we only need to consider the size of the domains associated with the edges. Finally, the *join* of two constraints is computed by reordering the trees so that the join variables occur at the top, merging the levels of the join variables in the same way as is done for splitting, and extending each leaf in the resulting tree with the cross-product of the corresponding subtrees of the original trees.

Constraint trees (and the way they are constructed) are close to the *hypercube* representation used in lifted belief propagation (Singla, Nath, & Domingos, 2010). However, for a given constraint, the constraint tree is typically more compact. The constraint tree of Figure 9 corresponds to a set of five hypercubes, one for each leaf. The hypercube representation does not exploit the fact that the first and second hypercube, for instance, share the part $\{x_1, x_2, x_3\}$. In the constraint tree, this is explicit, which makes it more compact.

We stress that GC-FOVE can use any extensionally complete constraint representation language. Constraint trees are just one such representation. Other representations can be more compact in some cases, but in the choice of a representation we need to consider also the tradeoff between compactness and ease of constraint processing. Consider a constraint graph, which is similar to our trees, but in which parent nodes can share child nodes. This representation is more compact than a constraint tree, but also requires more complicated constraint handling operations. For instance, consider splitting, in which we might need to split a child node for one parent but not for the others. Such operations become more complicated on graphs, while they are trivial on trees.

## 7. Experiments

Using an extensionally complete constraint language, we can capture more symmetries in the model, which potentially offers the ability to perform more operations at a lifted level. However, this comes at a cost, as manipulating more expressive constraints is more computationally demanding. We hypothesize that the ability to perform fewer computations by capturing more symmetries will far outweigh this cost in typical inference tasks. In this section, we compare the performances of C-FOVE and GC-FOVE$^{\text{TREES}}$ (GC-FOVE using the tree representation from Section 6) to empirically validate this hypothesis. In particular, we study how the performances vary as a function of two parameters: (i) the *domain size*, and (ii) the *amount of evidence*. We also empirically study whether GC-FOVE$^{\text{TREES}}$ can solve inference tasks that are beyond the reach of C-FOVE.

Throughout this section, GC-FOVE stands for GC-FOVE$^{\text{TREES}}$.

### 7.1 Methodology and Datasets

We compare C-FOVE and GC-FOVE on several inference tasks with synthetic and real-world data. We use the version of C-FOVE extended with general parfactor multiplication (de Salvo Braz, 2007).[3] For implementing GC-FOVE, we started from the publicly available C-FOVE code (Milch, 2008), so the implementations are maximally comparable.[4] In all experiments, the undirected model has parfactors whose constraints are all representable

---

3. This allows C-FOVE to handle some tasks in an entirely lifted way, where otherwise it would have to resort to grounding, e.g., on the *social network* domain (Jha et al., 2010).
4. GC-FOVE is available from `http://dtai.cs.kuleuven.be/ml/systems/gc-fove`.

by C-FOVE. Thus, GC-FOVE has no initial advantage, which makes the comparison conservative.

In each experiment we compute the marginal probability of a query randvar given some evidence. The query randvar is selected at random from the non-observed atoms. The evidence is generated by randomly selecting randvars of a particular predicate and giving them a value chosen randomly and uniformly from their domain. All the reported results are averaged over multiple runs for different query and evidence sets.

### 7.1.1 Experiments with Synthetic Data

In terms of synthetic data, we evaluate our algorithm on three standard benchmark problems. The first domain is called *workshop attributes* (Milch et al., 2008). Here, $m$ different attributes (e.g., topic, date, etc.) describe the workshop, and a corresponding factor for each attribute shows the dependency between the attendance of each person and the attribute. The theory contains the following parfactors.

$$\phi_1(Attends(X), Attr_1)$$
$$...$$
$$\phi_m(Attends(X), Attr_m)$$
$$\phi_{m+1}(Attends(X), Series)$$

The second domain is called *competing workshops* (Milch et al., 2008). It models the fact that people are more likely to attend a workshop if it is on a "hot topic" and that the number of attendees influences whether the workshop becomes a series. The theory contains the following parfactors.

$$\phi_1(Attends(X), Hot(Y))$$
$$\phi_2(Attends(X), Series)$$

In our experiments on both of the above domains, the query variable is *Series*, and all evidence randvars are of the form *Attends(x)*.

The third domain is called *social network* (Jha et al., 2010) and it models people's smoking habits, their chance of having asthma, and the dependence of a persons habits and diseases on their friendships. The theory contains the following parfactors.

$$\phi_1(Smokes(X))$$
$$\phi_2(Asthma(X))$$
$$\phi_3(Friends(X, Y))$$
$$\phi_4(Asthma(X), Smokes(X))$$
$$\phi_5(Asthma(X), Friends(X, Y), Smokes(Y))$$

In this domain, the evidence randvars will be a mix of randvars of the form *Smokes(x)* or *Asthma(x)*, and the query randvar can be any randvar that is unobserved.

### 7.1.2 Experiments with Real-World Data

We also used two other datasets from the field of statistical relational learning. The first, *WebKB* (Craven & Slattery, 1997), contains data about more than 1200 webpages, including their class (e.g., "course page"), textual content (set of words), and the hyperlinks between the pages. The model consists of multiple parfactors, stating for instance how the classes of two linked pages depend on each other. Our inference task concerns *link prediction*. Here, the class information is observed for a subset of all pages and the task is to compute the probability of having a hyperlink between a pair of pages. We use one Pageclass predicate in the model for each run, and average the runtime over multiple runs for each class. We used the following set of parfactors.

$$\phi_1(Pageclass(P))$$
$$\phi_2(Pageclass(P), HasWord(P, W))$$
$$\phi_3(Pageclass(P_1), Link(P_1, P_2), Pageclass(P_2))$$

The second dataset, *Yeast* (Davis, Burnside, de Castro Dutra, Page, & Costa, 2005), contains data about more than 7800 yeast genes, their functions and locations, and the interactions between these genes. The model and task are similar to those in *WebKB* (gene functions correspond to page classes, gene-to-gene interactions to hyperlinks). In this task, we observe the *function* information for a subset of all genes and query the existence of an *interaction* between two genes. Similar to WebKB, we also use one function in the model in each run and average the results over multiple runs. Here, we used the following set of parfactors.

$$\phi_1(Function(G))$$
$$\phi_2(Location(G, L))$$
$$\phi_3(Function(G), Location(G, L))$$
$$\phi_4(Function(G_1), Interaction(G_1, G_2), Function(G_2))$$

**Motivation for evidence randvars.** In all experiments, evidence randvars correspond to atoms of a unary predicate; we call them "unary randvars". This is done on purpose because introducing evidence randomly for binary randvars, e.g., randvars of the type $P(X, Y)$, can quickly break so many symmetries that lifted inference is not possible anymore. In fact, there are recent theoretical results that show that lifted inference in the presence of arbitrary evidence on binary randvars is simply not possible. This limitation is not unique to our approach, but is true of any possible exact lifted inference approach (Van den Broeck & Davis, 2012). Because of this, random insertion of evidence on binary randvars can quickly cause any lifted inference algorithm to resort to ground inference, which would blur the distinction between C-FOVE, GC-FOVE, and ground inference. We avoid this by placing evidence only on unary randvars.

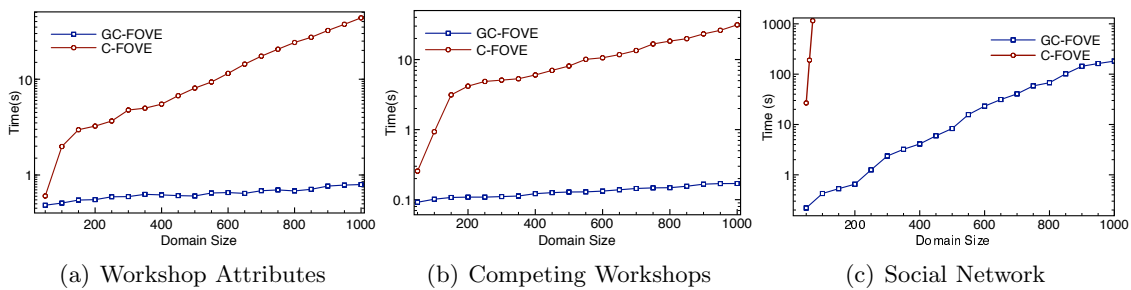

(a) Workshop Attributes    (b) Competing Workshops    (c) Social Network

Figure 10: Performance on synthetic data for varying domain sizes and proportion of observed randvars fixed at 20%. Y-axis (runtime) is drawn in log scale.

## 7.2 Influence of the Domain Size

In the first set of experiments, we use the synthetic datasets to measure the effect of domain size (number of objects) on runtime. We vary the domain size from 50 to 1000 objects while holding the proportion of observed randvars (relative to the number of observable randvars) constant at 20%. Figures 10(a) through 10(c) show the performance on all three synthetic datasets. On all three models, GC-FOVE outperforms C-FOVE on all domain sizes. As the number of objects in the domain increases, the runtimes increase for both algorithms. GC-FOVE's runtime increases at a much lower rate than C-FOVE's on all three models. On the first two tasks, GC-FOVE is between one and two orders of magnitude faster than C-GOVE, for the largest domain sizes. On the social network domain, the difference in performance becomes more striking: C-FOVE cannot handle domain sizes of 100 objects or more, while GC-FOVE handles the largest domain (1000 objects) in about 200 seconds. The improvement in performance arises as GC-FOVE better preserves the symmetries present in the model by treating all indistinguishable elements, observed or not, as a single unit.

The gain is more pronounced for larger domains. C-FOVE makes a separate partition (and a separate evidence factor) for each observed randvar, thus, with a fixed evidence ratio, the number of partitions induced by C-FOVE grows linearly with the domain size. Moreover, it has a costly elimination operation for each partition. In contrast, GC-FOVE, which employs lifted absorption, keeps the model at a higher granularity by grouping the observations and handles whole groups of observations with a single lifted operation.

## 7.3 Influence of the Amount of Evidence

In the second set of experiments, we measure the effect of the proportion of observed randvars on runtime, using the synthetic datasets. We fix the domain size, and vary the percentage of observed randvars from 0% to 100%. Note that this is a percentage of all "observable" randvars (e.g., all randvars of the form *Smokes(x)*), not of all randvars of any type (so 100% does not mean there are no unobserved variables left). Figures 11(a) through 11(c) show the performance on all three synthetic domains with domain size of 1000 objects. To better demonstrate C-FOVE's behavior on the social networks domain, Figure 11(d) shows the performances on a domain with only 25 objects. Both algorithms display similar trends across the three domains. Without evidence, GC-FOVE is comparable

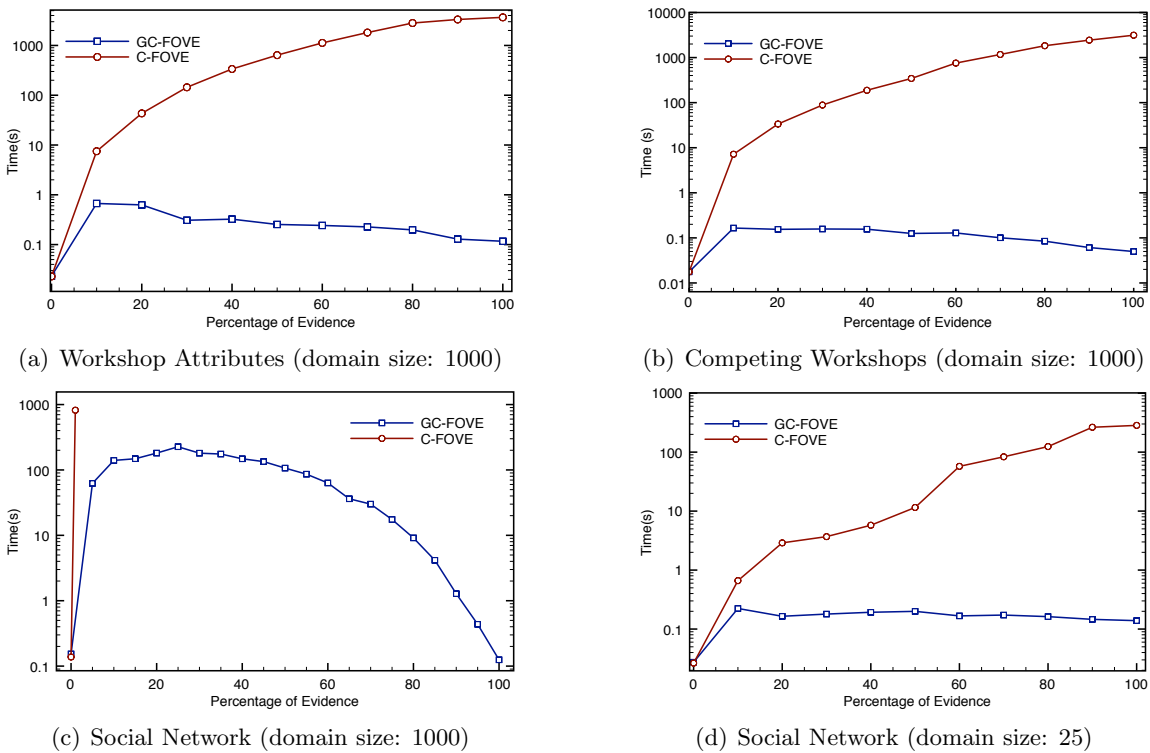

(a) Workshop Attributes (domain size: 1000)

(b) Competing Workshops (domain size: 1000)

(c) Social Network (domain size: 1000)

(d) Social Network (domain size: 25)

Figure 11: Performance on synthetic data with varying amounts of evidence and a fixed domain size. The Y-axis (runtime) is drawn in log scale.

to C-FOVE. This is the best scenario for C-FOVE as (i) the initial model only contains (in)equality constraints, and (ii) there is no evidence, so no symmetries are broken when the inference operators are applied. In this case, the only difference in runtime between the two algorithms is the overhead associated with constraint processing, which is almost negligible. As the proportion of observations increases, and the symmetries between the objects are broken, GC-FOVE maintains a much coarser grouping, and so performs inference much more efficiently, than C-FOVE. In all domains, C-FOVE's runtime increases dramatically with an increase in the percentage of observations. As more evidence is added, C-FOVE induces more partitions, which results in finer groupings of objects and leaves fewer opportunities for lifting. GC-FOVE performs significantly better in comparison, due to coarser grouping of observations and employing absorption for their elimination from the model. GC-FOVE's runtime experiences a bump as the initial set of evidence is added, but then levels out or gradually decreases (the more evidence, the more randvars are efficiently eliminated by absorption). GC-FOVE consistently finishes in under 200 seconds, regardless of the setting. In contrast, on the *social network* domain (Figure 11(c)) C-FOVE cannot handle portions of evidence greater than 1% (it runs out of memory on machine configured with 30GB of memory).

These results confirm that both the coarser groupings and the use of lifted absorption contribute to the much better performance of GC-FOVE.

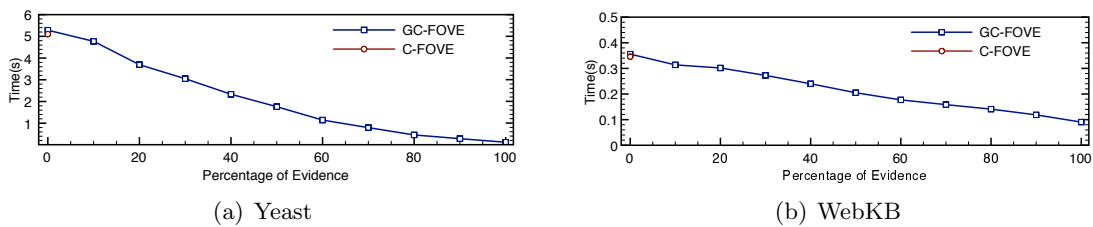

(a) Yeast         (b) WebKB

Figure 12: Performance on real-world data with varying amounts of evidence. The Y-axis (runtime) is drawn in log scale. C-FOVE only ran to completion for the zero-evidence experiments.

## 7.4 Performance on Real-World Data

In the final set of experiments, we compared the algorithms on the two real-world datasets, *WebKB* and *Yeast*. On both datasets, we varied the percentage of observed page classes or functions from 0% to 100% in steps of 10%. Figures 12(a) and 12(b) illustrate the results. C-FOVE could solve only the zero-evidence problems in these experiments; for the other cases, it typically ran out of memory after up to an hour of computation time on a machine configured with 30 GB of memory. Its failure is primarily due to the large number of observations, which often forces it to resort to inference at the ground level for a large number of objects. GC-FOVE, on the other hand, runs successfully for all experimental conditions. Furthermore, GC-FOVE can consistently solve the problems in a few seconds. As on the synthetic data, GC-FOVE's performance improves with increasing number of observations. In these cases more randvars can be eliminated through absorption, instead of the more expensive operations of multiplication and summation.

## 8. Conclusions

Constraints play a crucial role in lifted probabilistic inference as they determine the degree of lifting that takes places. Surprisingly, most lifted inference algorithms use the same class of constraints based on pairwise (in)equalities (Poole, 2003; de Salvo Braz et al., 2005; Milch et al., 2008; Jha et al., 2010; Kisynski & Poole, 2009b; Van den Broeck et al., 2011); the main exception is the work on approximate inference using lifted belief propagation (Singla & Domingos, 2008). In this paper we have shown that this class of constraints is overly restrictive. We proposed using *extensionally complete* constraint languages, which can capture more symmetries among the objects and allow for more operations to occur on a lifted level. We defined the relevant constraint handling operations (e.g., splitting and normalization) to work with extensionally complete constraint languages and implemented them for performing lifted variable elimination. We made use of constraint trees to efficiently represent and manipulate the constraints. We empirically evaluated our system on several domains. Our approach resulted in up to three orders of magnitude improvement in runtime, as compared to C-FOVE. Furthermore, GC-FOVE can solve several tasks that are intractable for C-FOVE.

Future work includes generalizing other lifted inference algorithms that currently use only inequality constraints, e.g., the works of Jha et al. (2010) and Van den Broeck et al. (2011), and further optimizing constraint handling. With respect to the latter, an interesting direction is the recent work of de Salvo Braz, Saadati, Bui, and OReilly (2012) that employs a logical representation for constraints, which is extensionally complete, and presents specialized constraint processing methods for this representation. Finally, it is possible to extend lifted absorption such that it works not only with evidence parfactors, but more generally with deterministic parfactors. This is another promising direction for future work.

**Acknowledgments**

Daan Fierens is supported by the Research Foundation of Flanders (FWO-Vlaanderen). Jesse Davis is partially supported by the Research Fund KULeuven (CREA/11/015 and OT/11/051), and EU FP7 Marie Curie Career Integration Grant (#294068). This work was funded by GOA/08/008 "Probabilistic Logic Learning" of the Research Fund KULeuven. The authors thank Maurice Bruynooghe and Guy Van den Broeck for interesting discussions and comments on this work and text. They also thank the reviewers for their constructive comments and very concrete suggestions to improve the article.

## Appendix A. Correctness Proof for Lifted Absorption

In this appendix, we prove the correctness of the novel lifted absorption operator. We begin by providing some lemmas.

Recall that a set of parfactors $G$ is a compact way of defining a set of factors $gr(G) = \{f | f \in gr(g) \land g \in G\}$ and the corresponding probability distribution

$$\mathcal{P}_G(\mathcal{A}) = \frac{1}{Z} \prod_{f \in gr(G)} \phi_f(\mathcal{A}_f).$$

Further, $G \sim G'$ means $G$ and $G'$ define the same probability distribution. Thus, formally:

$$G \sim G' \Leftrightarrow \mathcal{P}_G(\mathcal{A}) = \mathcal{P}_{G'}(\mathcal{A}) \Leftrightarrow \frac{1}{Z} \prod_{f \in gr(G)} \phi_f(\mathcal{A}_f) = \frac{1}{Z'} \prod_{f \in gr(G')} \phi_f(\mathcal{A}_f).$$

The following lemmas are easily proven by applying the above definition and keeping in mind that $gr(G \cup G') = gr(G) \cup gr(G')$.

**Lemma 1** *For all models $G, G', G''$: $G' \sim G'' \Rightarrow G \cup G' \sim G \cup G''$.*

**Lemma 2** *Given a factor $f = \phi(A_1, A_2, \ldots, A_n)$ and an evidence factor $f_E = \phi_E(A_1)$ with $\phi_E(a_1) = 1$ if $a_1 = o$ (the observed value) and $\phi_E(a_1) = 0$ otherwise, $\{f, f_E\} \sim \{f', f_E\}$ with $f' = \phi'(A_2, \ldots, A_n)$ and $\phi'(a_2, \ldots, a_n) = \phi(o, a_2, \ldots, a_n)$.*

**Lemma 3** *A model that consists of $m$ identical factors, $G = \{\phi(A_1, \ldots, A_n)\}_{i=1}^{m}$, is equivalent to a model with a single factor $G' = \{\phi'(A_1, \ldots, A_n)\}$ where $\phi'(a_1, \ldots, a_n) = \phi(a_1, \ldots, a_n)^m$.*

We now prove that the ABSORB operator is correct, i.e., its postconditions hold, given the preconditions.

**Theorem 1** *Given a model $G$, a parfactor $g \in G$ and an evidence parfactor $g_E$, if the preconditions of the* ABSORB *operator are fulfilled, then*

$$G \cup \{g_E\} \sim G \setminus \{g\} \cup \{\text{ABSORB}(g, A_i, g_E), g_E\}.$$

**Proof:** With $G' = G \setminus \{g\}$, we can rewrite the above equivalence as

$$G' \cup \{g, g_E\} \sim G' \cup \{\text{ABSORB}(g, A_i, g_E), g_E\}.$$

Because of Lemma 1, it suffices to prove

$$\{g, g_E\} \sim \{\text{ABSORB}(g, A_i, g_E), g_E\}.$$

Let $g = \phi(\mathcal{A})|C$ with $\mathcal{A} = \{A_1(\mathbf{X}_1), \ldots, A_k(\mathbf{X}_k)\}$, let $g_E = \phi_E(P(\mathbf{X}))|C_E$, and let $L = logvar(\mathcal{A})$ (non-counted logvars in $\mathcal{A}$), $\mathbf{X}^{excl} = \mathbf{X} \setminus logvar(\mathcal{A} \setminus \{A_i\})$ (logvars occurring exclusively in $A_i$) and $L' = logvar(\mathcal{A}) \setminus \mathbf{X}^{excl}$ (non-counted logvars occurring (also) outside $A_i$). The operator returns a parfactor of the form $\phi'(\mathcal{A}')|C'$ where $\mathcal{A}' = \{A_2, \ldots, A_k\}$ and $C' = \pi_{logvar(C) \setminus \mathbf{X}^{excl}}(C)$ (see the operator definition). We need to prove that $\phi'$ is such that the above equivalence holds. For ease of exposition, from now we assume that the atom or counting formula that is to be absorbed ($A_i$ in the operator's input) is $A_1$. We first consider the case where $A_1$ is an atom $P(\mathbf{X}_1)$, and then the case where $A_1$ is a counting formula $\#_X[P(\mathbf{X}_1)]$.

**Absorption for atoms.** In this case, we have $\phi'(a_2, \ldots, a_k) = \phi(o, a_2, \ldots, a_k)^r$ with $r = \text{COUNT}_{\mathbf{X}^{excl}|L'}(C)$ (see the operator definition, observing that $\mathbf{X}^{nce} = \mathbf{X}^{excl}$).

By definition, $gr(g) = \{\phi(P(\mathbf{x}_1), A_2(\mathbf{x}_2), \ldots, A_k(\mathbf{x}_k))\}_{\mathbf{l} \in \pi_L(C)}$, with $\mathbf{x}_i = \pi_{\mathbf{X}_i}(\mathbf{l})$. Precondition 1 guarantees that for each $\phi(P(\mathbf{x}_1), A_2(\mathbf{x}_2), \ldots, A_k(\mathbf{x}_k))$, there exists an evidence factor $\phi_E(P(\mathbf{x}_1))$ in $gr(g_E)$. By Lemma 2, we can therefore rewrite each factor in $gr(g)$ into the form $\phi^*(A_2(\mathbf{x}_2), \ldots)$, with $\phi^*(a_2, \ldots a_k) = \phi(o, a_2, \ldots, a_k)$, with $o$ the observed value for $P(\mathbf{x}_1)$.

The potential function $\phi^*$ is the same for all factors, since there is only one observed value $o$ for the whole evidence parfactor. Therefore, any two factors $\phi(P(\mathbf{x}_1), A_2(\mathbf{x}_2), \ldots, A_k(\mathbf{x}_k))$ and $\phi(P(\mathbf{x}'_1), A_2(\mathbf{x}_2), \ldots, A_k(\mathbf{x}_k))$ that differ only in their first argument are rewritten to the same factor. Because of Precondition 2, the number of factors rewritten to the same factor is constant and equals $\text{COUNT}_{\mathbf{X}^{excl}|L'}(C) = r$. By Lemma 3, each set of identical factors can therefore be replaced by a single factor with potential function

$$\phi'(a_2, \ldots, a_k) = \phi^*(a_2, \ldots, a_k)^r = \phi(o, a_2, \ldots, a_k)^r,$$

which is exactly how $\phi'$ is defined by the operator.

**Absorption for counting formulas.** In this case, we have

$$\phi'(a_2, \ldots, a_k) = \phi(e, a_2, \ldots, a_k)^r$$

and $r = \text{COUNT}_{\mathbf{X}^{nce}|L'}(C)$ with $\mathbf{X}^{nce} = \mathbf{X}^{excl} \setminus \{X\}$ (see the operator definition).

We define $\mathbf{X}_1' = \mathbf{X}_1 \setminus \{X\}$, and use $(\mathbf{x}_1', X)$ to denote $\mathbf{X}_1$ with all logvars instantiated except the counted logvar $X$. Now, by definition,

$$gr(g) = \{\phi(\#_{X \in C_1}[P(\mathbf{x}_1', X)], A_2(\mathbf{x}_2), \ldots, A_k(\mathbf{x}_k))\}_{\mathbf{l} \in \pi_L(C)},$$

with $\mathbf{x}_i = \pi_{\mathbf{X}_i}(\mathbf{l})$, $\mathbf{x}_1' = \pi_{\mathbf{X}_1'}(\mathbf{l})$ and $C_1 = \pi_X(\sigma_{L=1}(C))$. Each $C_1$ is of the form $\{x_1, \ldots, x_n\}$, where $n = \text{Count}_{X|L}(C)$ ($n$ exists because PCRV's are by definition count-normalized).

We show correctness of the operator in this case by showing that for each factor $f$ in $gr(g)$, the evidence parfactor $g_E$ can be rewritten to contain an evidence factor that has the same CRV as $f$, such that the same reasoning as above can be applied on $f$.

Precondition 1 guarantees that for each factor

$$f = \phi(\#_{X \in \{x_1, \ldots, x_n\}}[P(\mathbf{x}_1', X)], A_2(\mathbf{x}_2), \ldots, A_k(\mathbf{x}_k)),$$

$gr(g_E)$ contains the group of evidence factors

$$E_f = \{\phi_E(P(\mathbf{x}_1', x_1)), \ldots, \phi_E(P(\mathbf{x}_1', x_n))\}.$$

We can multiply all factors in $E_f$ into

$$\phi_E'(P(\mathbf{x}_1, x_1'), \ldots, P(\mathbf{x}_1, x_n')),$$

with $\phi_E'(o, o, \ldots, o) = 1$ and $\phi_E'(.) = 0$ elsewhere, and then rewrite this as

$$f_E = \phi_E^*(\#_{X \in \{x_1, \ldots, x_n\}}[P(\mathbf{x}_1', X)]),$$

where $\phi_E^*$ is such that (i) $\phi_E^*(e) = 1$, for $e$ the histogram with $e(o) = n$ and $e(.) = 0$ elsewhere, and (ii) $\phi_E^*(e') = 0$ for $e' \neq e$.

Having formed $f_E$, we can rewrite $f$ into the form $\phi'(A_2(\mathbf{x}_2), \ldots)$, with $\phi'(a_2, \ldots a_k) = \phi(e, a_2, \ldots, a_k)^r$ and $r = \text{Count}_{\mathbf{X}^{nce}|L'}(C)$, with the same argumentation as for regular atoms. After this, we can replace $f_E$ with the equivalent $E_f$, thus restoring $g_E$. Repeating this for each $f$ preserves equivalence and eventually yields the model that the operator returns. □

## Appendix B. Computational Complexity of Lifted Absorption

Applying lifted absorption on a parfactor $g = \phi(\mathcal{A})|C$, has complexity $O(|C|) + O(Size(\phi) \cdot \log |C|)$, where $|C|$ is the cardinality (number of tuples) of the constraint $C$, and $Size(\phi)$ equals the product of range sizes of the arguments $\mathcal{A}$, i.e., $Size(\phi) = \prod_{A_i \in \mathcal{A}} |range(A_i)|$. The first term in the complexity, $O(|C|)$, arises because absorption involves a projection of the constraint $C$, which in the worst case (with an extensional representation) has complexity $O(|C|)$. The second term, $O(Size(\phi) \cdot \log |C|)$, is the complexity of computing the new potential function, which involves manipulating $\phi$, which has $Size(\phi)$ entries (in a tabular representation), and exponentiating it, which has complexity $O(\log |C|)$.